\documentclass[letterpaper,journal]{IEEEtran}

\usepackage{amsmath,amsfonts}
\usepackage{graphicx}
\usepackage{cite}
\usepackage{array}
\usepackage{textcomp}
\usepackage{url}
\usepackage{algorithm}
\usepackage{hyperref}
\usepackage{amssymb}
\usepackage{booktabs}
\usepackage{subcaption}
\usepackage{algpseudocode}

\begin{document}

\title{Connectivity-Aware Exploration of Robotic Grasp Spaces}

\author{Maksim Kazanskii$^{*}$%
\thanks{Independent Researcher. $^{*}$ mkazanskii@gmail.com.}}

\maketitle

\begin{abstract}
Robotic grasping typically focuses on identifying successful
actions among candidate grasp poses, while the collective
organization of successful actions remains less explored.
We investigate the multiscale connectivity of viable grasps
in $SE(3)$ and its value for exploration and cross-object
transfer. Using the ACRONYM dataset, we demonstrate
heterogeneous, reproducible, and object-dependent connectivity
patterns. We introduce connectivity-aware sampling that
prioritizes potential bridges, structural frontiers,
boundary extensions, and geometric novelty. In controlled
reconstruction experiments, this strategy recovers viable
grasp-space connectivity more efficiently than random
sampling and farthest-point sampling, achieving median
sample-efficiency gains of up to $4.76\times$.
When grasp success is initially unknown, proximity to
both successful and failed observations improves grasp
discovery. Explicit connectivity modeling provides
little additional benefit in this setting. We further
introduce a cross-object exploration framework combining
connectivity-selected reference prototypes, local
graph-structural retrieval, and online success--failure
feedback. On held-out objects, structural retrieval
outperforms geometric retrieval by approximately
1.5--6.0 successful grasps per 100 queries across
evaluated reference-scaling conditions. Together, these
findings demonstrate the value of grasp-space connectivity
for structural reconstruction and cross-object transfer.
\end{abstract}


\section{Introduction}

\IEEEPARstart{R}{obotic} manipulation requires selecting
effective actions from large, continuous spaces of possible
configurations. In robotic grasping, this typically involves
selecting a grasp pose $g\in SE(3)$ that can successfully
grasp an object $O$. Much of modern grasping research focuses
on generating candidate grasps and evaluating their quality
or likelihood of success
~\cite{mahler2017dexnet,ten2017gpd,liang2019pointnetgpd,
sundermeyer2021contact}.
For probabilistic grasp-quality models, this can be expressed as
\begin{equation}
Q(O,g)\approx P(\mathrm{success}\mid O,g).
\end{equation}

Although effective for evaluating individual actions, this
formulation does not explicitly characterize how successful
grasps are organized collectively. Previous work has
investigated structured grasp spaces, continuous manifolds,
and topological relationships
~\cite{zarubin2012topological,pokorny2013grasp,
pokorny2014grasp,hang2016evolution,hager2021graspme}.
Here, we examine the connectivity of the viable subset
of grasp space and its potential value for exploration.
Two objects may have similar fractions of successful
candidate grasps but different spatial organizations:
successful grasps may form a few broad connected regions
or many separated components. Such differences may affect
how efficiently an agent can reconstruct viable grasp-space
structure and discover successful actions.

For an object $O$, we define the viable grasp set as
\begin{equation}
\mathcal{G}^{+}(O)=
\left\{
g\in SE(3):
g\text{ successfully grasps }O
\right\}.
\end{equation}
We characterize its multiscale connectivity using a
normalized $SE(3)$ distance. At neighborhood radius
$\epsilon$, two viable grasps are directly connected
when their distance does not exceed $\epsilon$.
Connected components are formed through sequences
of such neighboring grasps. Varying $\epsilon$ reveals
how local regions merge into larger components,
analogous to zeroth-dimensional topological filtrations
~\cite{edelsbrunner2010computational,mahler2018caging}.
This representation captures aspects of viable grasp-space
organization beyond the number of successful candidates.

Using simulated parallel-jaw grasps from the ACRONYM
dataset~\cite{eppner2021acronym}, we first examine
whether multiscale connectivity is reproducible across
independent samples. We find that within-object
connectivity signatures become increasingly consistent
with sample size, while between-object differences
remain larger. These observations suggest that the
signatures capture object-dependent structure beyond
finite-sample variability.

We then investigate whether connectivity can guide
structural reconstruction. Random sampling does not
account for observed structure, while farthest-point
sampling (FPS) prioritizes geometric coverage
~\cite{gonzalez1985clustering,qi2017pointnetplusplus}.
We introduce a connectivity-aware strategy that
prioritizes potential bridges between components,
structural frontiers, boundary extensions, and
geometric novelty. In a controlled setting where
candidate grasps are known to be viable, this
strategy reconstructs reference connectivity
more efficiently than random sampling and FPS,
achieving median sample-efficiency gains of up
to $4.76\times$ over random sampling.

We next consider hidden grasp viability, where
candidate poses are known but their outcomes
are revealed only through queries. Observations
acquired from one candidate pool are used to rank
untested grasps in a disjoint pool. We find that
using proximity to both successful and failed
observations improves discovery compared with
using successful observations alone, while
explicit connectivity modeling provides little
additional benefit. Thus, the advantages of
connectivity for structural reconstruction do
not necessarily extend to individual grasp-success
prediction.

Finally, we examine cross-object transfer.
We construct reference memories by selecting
representative grasps from connected viable
regions of previously explored objects.
For unseen objects, local structural signatures
computed from unlabeled candidate-pose graphs
enable retrieval of reference experience without
requiring known target grasp outcomes. Retrieved
information is combined with success--failure
feedback acquired during exploration. We compare
connectivity-based prototype selection with random
and geometric selection, and structural retrieval
with pose-based retrieval using matched reference
prototypes. Experiments on held-out objects
demonstrate improved grasp discovery through
structural retrieval. Reference and evaluation
objects are disjoint, although object categories
are not explicitly separated.

The main contributions are:
\begin{itemize}

\item We characterize the multiscale connectivity
of viable grasps in $SE(3)$ and demonstrate
reproducible, object-dependent structural patterns.

\item We introduce connectivity-aware sampling
that improves structural reconstruction efficiency
by up to $4.76\times$ relative to random sampling
at matched reconstruction quality.

\item We investigate hidden-viability exploration
and show that local success--failure geometry
captures most of the predictive benefit, with
limited additional improvement from explicit
connectivity.

\item We introduce a cross-object exploration
framework combining connectivity-selected
reference memories, local structural retrieval,
and online feedback, demonstrating improved
grasp discovery on held-out objects.

\end{itemize}

\section{Related Work}
\label{sec:related_work}

\subsection{Robotic Grasp Generation and Evaluation}

Robotic grasping is commonly formulated as the generation and evaluation
of candidate actions. Classical approaches characterize grasp quality
through geometric and physical criteria such as force closure, grasp
wrench space, and robustness to uncertainty
~\cite{nguyen1988constructing,ferrari1992planning,prattichizzo2016grasping}.
These formulations provide principled criteria for evaluating individual
grasps, but typically do not characterize the organization of the complete
set of viable grasp configurations.

Data-driven methods instead learn mappings from sensory observations and
candidate configurations to grasp quality or success. Early data-driven
approaches demonstrated that grasp detection could be learned directly
from visual observations~\cite{lenz2015deep,redmon2015real}. More recent
methods increasingly operate in full 6-DoF pose spaces. GPD generates and
scores candidate 6-DoF grasps from point clouds
~\cite{ten2017gpd}, while PointNetGPD evaluates grasp configurations
directly from local 3D point clouds~\cite{liang2019pointnetgpd}.
Contact-GraspNet predicts 6-DoF grasps directly from scene point clouds
~\cite{sundermeyer2021contact}.

Large-scale synthetic datasets have enabled increasingly dense sampling of
robotic grasp spaces. Dex-Net 2.0 used millions of synthetic grasp examples
for grasp-quality learning~\cite{mahler2017dexnet}, while GraspNet-1Billion
introduced a large-scale benchmark for 6-DoF grasp pose detection
~\cite{fang2020graspnet}. ACRONYM provides 17.7 million simulated
parallel-jaw grasps over 8,872 objects with physics-based grasp outcomes
~\cite{eppner2021acronym}. Such datasets make it possible to study
properties of grasp spaces beyond the quality of individual actions.

These approaches primarily address the generation, quality, or ranking of
individual grasp actions. In contrast, we study the organization of the
viable grasp set itself, focusing on how successful actions are distributed
and connected across multiple spatial scales.

\subsection{Geometry and Manifolds of Grasp Spaces}

The representation of grasps as elements of structured geometric spaces
has been explored in several forms. Pokorny et al.\ introduced
\emph{Grasp Moduli Spaces}, jointly representing objects and grasps in a
continuous metric space and studying how stable grasps behave under
continuous deformations of object shape and grasp configuration
~\cite{pokorny2013grasp}. Their subsequent work used this representation
together with shape descriptors to reason about grasp transfer and
optimization across objects~\cite{pokorny2014grasp}.

Other work has represented sets of possible grasps as structured spaces
rather than as isolated candidates. Hang et al.\ organized multi-fingered
grasps according to fingertip configurations~\cite{hang2014hierarchical}
and later learned continuous grasping manifolds from accumulated robot
experience to support grasp planning on new objects
~\cite{hang2016evolution}. GraspME estimates continuous grasp manifolds representing sets of possible end-effector positions, allowing them to be used as goal regions for motion optimization~\cite{hager2021graspme}. Zarubin et al. developed a topological representation of stable grasps to support grasp transfer across different hand kinematics~\cite{zarubin2012topological}.

Our work differs from these approaches in both the structure being studied
and how that structure is used. Rather than learning a continuous grasp
manifold, we characterize the multiscale connectivity of the successful
subset of a 6-DoF grasp space. We then investigate whether this connectivity
can guide the selection of informative grasp candidates. For cross-object
transfer, we construct reference memories from representative successful
grasps selected within connected regions of previously explored objects.
Local structural signatures are used to retrieve reference prototypes
for a new object, and the retrieved information is combined with online
success--failure feedback to rank untested grasp candidates.

\subsection{Topology, Connectivity, and Caging}

Topological and connectivity-based reasoning has a long history in robotic
manipulation, particularly in caging. In robotic caging, an object need
not be rigidly immobilized. Instead, the robot constrains its motion
to a bounded region from which it cannot escape. Caging methods
therefore study the connectivity of the object's configuration space
and the existence of escape paths
~\cite{rimon1999caging,pipattanasomporn2006two}. Subsequent approaches
extended these ideas to more general manipulation settings and emphasized
the role of connected components and configuration-space structure in
determining whether an object is constrained
~\cite{makita2008caging,rodriguez2012caging}.

More recently, topological tools have been used to characterize such
structure across scales. Mahler et al.\ applied persistent homology to
energy-bounded planar caging, using the evolution of topological structure
to identify configurations from which escape requires overcoming an
energy barrier~\cite{mahler2018caging}. More generally, zeroth-dimensional
persistent homology tracks the appearance and merging of connected
components as a filtration parameter varies
~\cite{edelsbrunner2010computational}. This provides a natural framework
for describing connectivity across multiple scales.

Our setting differs in both the underlying space and the role of topology.
We do not analyze the free configuration space of an object inside a cage
or synthesize caging configurations. Instead, our points are successful
parallel-jaw grasp poses in $SE(3)$, and we study how the connectivity of
this empirically sampled viable-action set changes with neighborhood scale.
We use zeroth-dimensional connectivity---the number of connected components
and the relative size of the largest component---as a multiscale
description of the viable grasp set. We do not compute higher-dimensional
topological features or use persistence diagrams. Instead, we investigate
whether this connectivity is reproducible, whether it can guide the
selection of new grasp observations, and whether the resulting exploration
experience can improve subsequent grasp discovery.
\subsection{Grasp-Space Summarization and Sampling}

Large grasp datasets are often highly redundant, motivating methods for
constructing compact representations. Hjelm et al.\ proposed sparse
summarization of high-dimensional continuous robotic grasping data using
a Bayesian non-parametric representation combining dimensionality
reduction and clustering~\cite{hjelm2013sparse}. Related work has also developed compact representations for organizing
grasp experience and transferring grasp knowledge across objects
~\cite{madry2012storing,hjelm2014}.

Prior work has also considered the explicit construction and exploration of
grasp spaces. Roa et al.\ constructed graspable and non-graspable regions
from sampled force-closure and non-force-closure grasps, with the goal of
efficiently generating the grasp space
~\cite{roa2008grasp}. Rolinat et al.\ learned a compact representation of
grasp space from a limited set of expert grasps using a variational
autoencoder and used the learned representation to generate new grasp
configurations~\cite{rolinat2021grasp}. These approaches focus on
constructing or generating feasible grasp configurations, whereas we study
the multiscale connectivity of successful grasps and use the connectivity
observed during exploration to guide the selection of new candidates.

A simple geometric strategy for constructing representative subsets is
farthest-point sampling (FPS), which iteratively selects the point farthest
from the currently selected set. FPS is widely used for geometric
subsampling; for example, PointNet++ uses iterative FPS to obtain
well-distributed centroids in point clouds~\cite{qi2017pointnetplusplus}.

Random sampling provides a baseline that does not use
geometric or connectivity information. In contrast, FPS
prioritizes geometric coverage, while our connectivity-aware
method uses the evolving structure of observed viable grasps
to identify informative candidates. Comparing these three
strategies in Experiment~2 tests whether connectivity-aware
sampling improves structural reconstruction beyond uniform
sampling and geometric coverage.

\subsection{Cross-Object Transfer and Structural Retrieval}

Transferring grasp knowledge across objects requires identifying
relationships between grasp configurations despite differences
in object geometry and pose distributions. Existing approaches
have explored transferring grasp experience through geometric
descriptors, object similarity, and learned representations
~\cite{madry2012storing,hjelm2014,pokorny2014grasp}.
These methods motivate the use of previously acquired grasp
experience to improve predictions on new objects.

A complementary direction is to represent local geometric
structure rather than relying exclusively on absolute
coordinates. Graph-based representations and local geometric
descriptors provide mechanisms for characterizing neighborhood
relationships in irregular spatial data
~\cite{qi2017pointnetplusplus,wang2019dynamic}.
Such representations suggest that local candidate distributions
may provide transferable information even when direct
cross-object pose correspondence is unavailable.

Our approach combines connectivity-informed reference selection
with structural retrieval. Successful reference grasps are
organized into connected regions, from which representative
prototypes are selected. Each prototype is associated with a
structural signature computed from the unlabeled candidate-pose
graph of its source object. For a new object, reference prototypes
are retrieved by similarity between these local signatures,
and previously observed target outcomes dynamically influence
their ranking. Unlike approaches that require matching
successful grasp regions across objects, this mechanism does
not assume that the connectivity of the target's viable grasp
set is known.

We evaluate the contributions of prototype selection and
retrieval representation separately, comparing random,
geometrically dispersed, and connectivity-selected memories,
as well as pose-based and structural retrieval under matched
reference-memory sizes.

\subsection{Active and Exploratory Grasping}

A complementary literature considers how robots should select actions
sequentially when grasp trials or observations are costly. Active learning
and Bayesian optimization provide general frameworks for choosing
informative evaluations rather than sampling uniformly
~\cite{settles2009active,shahriari2016bayesian}. These ideas are
particularly relevant to robotic manipulation, where physical interaction
can make each additional observation expensive.

In robotic grasping, several approaches have therefore treated grasp
acquisition as a sequential learning problem. Kroemer et al.\ combined
active learning with reactive control to improve grasp learning from
experience~\cite{kroemer2010active}. Mahler et al.\ used multi-armed
bandit methods to allocate simulation effort toward promising grasp
candidates~\cite{mahler2016dexnet}. More recently, Danielczuk et al.
introduced \emph{Exploratory Grasping}, in which a robot discovers reliable
grasps for unknown objects through repeated grasping and interaction
~\cite{danielczuk2021exploratory}. Their BORGES method uses experience
collected online to identify high-performing grasps efficiently. This work
is particularly related to ours in treating grasp acquisition as a
sequential exploration problem rather than as a static prediction task.

Experience can also be transferred across objects. Prior work has explored cross-object grasp transfer using semantic object attributes and geometric representations of objects and grasps~\cite{hjelm2014,pokorny2014grasp}. These approaches demonstrate that grasp knowledge acquired from previously encountered objects can inform grasping on new objects.

Our approach differs in both within-object exploration and cross-object
transfer. Within an object, we use the evolving multiscale connectivity of
observed successful grasps to select structurally informative candidates.
Across objects, we construct reference memories from representative
successful grasps selected within connected regions of previously
explored objects. For a new object, local structural signatures enable
retrieval of reference prototypes, while online success--failure feedback
informs candidate ranking.

\section{Methods}
\label{sec:methods}

\subsection{Problem Formulation}

We represent a parallel-jaw grasp as a rigid transformation
\begin{equation}
    g = (\mathbf{p}, R) \in SE(3),
\end{equation}
where $\mathbf{p}\in\mathbb{R}^{3}$ denotes the gripper position and
$R\in SO(3)$ its orientation. For an object $O$, we define the viable
grasp set as
\begin{equation}
    \mathcal{G}^{+}(O)
    =
    \left\{
        g\in SE(3):
        g\text{ successfully grasps }O
    \right\}.
\end{equation}

Rather than treating the elements of \(\mathcal G^+(O)\) only as individual positive examples, we study their organization in action space. In
particular, we characterize the connectivity of the viable set over a
range of neighborhood scales and ask whether this structure can be
recovered efficiently from partial observations.

We first test whether independently sampled subsets of the same object's viable grasp set exhibit consistent multiscale structure. We then ask how efficiently this structure can be reconstructed from partial observations of a finite viable reference set. This controlled setting assumes known viability, isolating structural reconstruction from the separate problem of discovering grasp success.

\subsection{Dataset and Grasp Representation}

We use the ACRONYM dataset~\cite{eppner2021acronym}, which contains
17.7 million simulated parallel-jaw grasps for 8,872 objects from 262
object categories. Each candidate grasp is represented by a homogeneous
$4\times4$ transformation matrix and is accompanied by a simulated grasp
outcome.

A grasp is considered viable when the
\texttt{object\_in\_gripper} label indicates successful retention of the
object. Objects are included in an experiment only when they contain at
least the required number of successful grasps.

\subsection{Normalized Distance in \texorpdfstring{$SE(3)$}{SE(3)}}
\label{sec:se3_metric}

To compare grasp poses, we define a dimensionless distance that combines
translation and rotation. For two grasps
$g_i=(\mathbf{p}_i,R_i)$ and $g_j=(\mathbf{p}_j,R_j)$, the translational
distance is
\begin{equation}
    d_p(i,j)
    =
    \left\|
        \mathbf{p}_i-\mathbf{p}_j
    \right\|_2.
\end{equation}

Rotational separation is measured using the geodesic distance on
$SO(3)$,
\begin{equation}
    d_R(i,j)
    =
    \arccos
    \left(
        \frac{
            \operatorname{tr}(R_i^\top R_j)-1
        }{2}
    \right).
\end{equation}

Because translation and rotation have different units and characteristic
scales, we normalize each term separately. Let
\begin{equation}
    s_p
    =
    \operatorname{median}_{i\neq j}
    d_p(i,j),
    \qquad
    s_R
    =
    \operatorname{median}_{i\neq j}
    d_R(i,j),
\end{equation}
where the medians are computed over nonzero pairwise distances within the
grasp set used to define the normalization for each object. The resulting
normalization constants $s_p$ and $s_R$ are therefore object-specific and
are held fixed when comparing different subsets of the same object's grasp
space. The combined distance is then
\begin{equation}
    d(g_i,g_j)
    =
    \sqrt{
        \left(
            \frac{d_p(i,j)}{s_p}
        \right)^2
        +
        \left(
            \frac{d_R(i,j)}{s_R}
        \right)^2
    }.
    \label{eq:se3_distance}
\end{equation}

This normalization gives translation and rotation comparable scale without
introducing a manually chosen conversion factor between linear and angular
displacement.

\subsection{Experimental Overview}
\label{sec:experimental_overview}

We evaluate viable grasp-space structure in four experiments that progressively move from structural characterization to exploration under increasingly limited information.
\par\vspace{0.7em}
\paragraph*{Experiment 1}
E1 tests whether multiscale connectivity is a reproducible and
object-specific property of viable grasp spaces. We randomly select 100
objects with at least 300 successful grasps. For each object and
\[
N\in\{10,20,50,100,200,300\},
\]
we draw five independent subsets of $N$ successful grasps, indexed by
$r=1,\ldots,5$, and compute their signatures $T_{i,N}^{(r)}$. We quantify
within-object convergence as
\[
E_i(N)=\frac{1}{10}\sum_{r<s}
D\!\left(T_{i,N}^{(r)},T_{i,N}^{(s)}\right),
\]
where $D$ is the structural distance between two connectivity signatures,
defined as the equally weighted mean absolute difference between their
$B(\epsilon)$ and $L(\epsilon)$ curves (Appendix~\ref{app:connectivity})
and the sum is over 10 pairs of replicates. For each object, the normalization of the $SE(3)$ distances is computed
once from the complete successful-grasp set and held fixed across all
sample sizes and replicates. This ensures that differences between
signatures reflect grasp sampling rather than changes in the distance
normalization itself. The same neighborhood-radius grid is used throughout. We report the mean $E_i(N)$ across objects as a function of $N$, with
95\% bootstrap confidence intervals across objects.

To test object specificity, at $N=300$ we compare distances between
independent samples of the same object,
$D(T_{i,300}^{(r)},T_{i,300}^{(s)})$, with distances between samples from
different objects,
$D(T_{i,300}^{(r)},T_{j,300}^{(s)})$, $i\neq j$.
Reproducibility and object specificity correspond respectively to
decreasing $E_i(N)$ with $N$ and smaller within-object than between-object
distances. Object-specific normalization removes differences in absolute grasp-space scale, so between-object comparisons reflect differences in relative multiscale connectivity rather than object size.
\par\vspace{0.7em}
\paragraph*{Experiment 2: Known-Viability Structural Reconstruction}

E2 asks a simple question: if the complete set of successful grasps is
available, can we recover its connectivity structure while examining only
a small subset of those grasps? All candidate grasps in E2 are known to be
successful. The challenge is therefore not to discover successful grasps,
but to select the successful grasps that are most informative about the
structure of the complete viable set.

We first identify objects containing at least 500 successful grasps and
randomly select 200 eligible objects. For each selected object, we then
randomly sample 500 successful grasps and treat this set as the reference
viable space. Pairwise normalized $SE(3)$ distances
are computed between all reference grasps, and the reference connectivity
signature $T_{\mathrm{ref}}=(\mathbf{B}_{\mathrm{ref}},
\mathbf{L}_{\mathrm{ref}})$ is evaluated at 15 neighborhood radii uniformly
spaced over $\epsilon\in[0.10,0.80]$. The complete reference signature is
used only for evaluation and is never provided to the sampling methods.
Each method instead constructs an ordered subset of at most 200 grasps and,
after each observation, we recompute its connectivity signature and measure
the structural error as follows:
\begin{equation}
\begin{split}
E(T_n,T_{\mathrm{ref}})
=
\frac{1}{2}\Big[
&\operatorname{MAE}(\mathbf B_n,\mathbf B_{\mathrm{ref}})\\
&+\operatorname{MAE}(\mathbf L_n,\mathbf L_{\mathrm{ref}})
\Big].
\end{split}
\end{equation}

We compare three selection strategies. \emph{Random} samples viable grasps
uniformly without replacement. \emph{Farthest-point sampling} (FPS) is a geometry-only baseline that
repeatedly selects the candidate maximizing its normalized $SE(3)$ distance,
defined in Section~\ref{sec:se3_metric}, to the nearest previously selected
grasp~\cite{gonzalez1985clustering}. The
\emph{connectivity-aware (CA)}  strategy uses only the geometry and connectivity
of the currently observed subset. The connectivity-aware sampler begins by selecting five grasps uniformly
at random. Subsequent grasps are selected using an acquisition score that combines
four criteria: 

(i) \emph{bridging}, favoring candidates that lie within
the neighborhood radius of multiple currently disconnected components;

(ii) \emph{frontier exploration}, favoring candidates that lie just
outside the neighborhood radius of an observed component and therefore
probe unexplored space beyond its current extent;

(iii) \emph{boundary
extension}, favoring candidates that lie within the neighborhood radius
of exactly one observed component and therefore extend that component
without connecting it to another; 

(iv) \emph{geometric novelty},
favoring candidates that are distant from previously selected grasps and
therefore discouraging repeated sampling within the same local region.

We prioritize bridge formation over frontier and boundary exploration,
while assigning a smaller weight to geometric novelty. The bridge,
frontier, boundary, and geometric-novelty terms are assigned weights of
$2.0$, $1.0$, $0.5$, and $0.20$, respectively. The weighting is heuristic,
reflecting the intended priority of resolving uncertain connectivity before
promoting broader geometric coverage.

We evaluate structural reconstruction efficiency by measuring how many
grasps each method must select to reach a fixed structural reconstruction
error. For each object, Random sampling is evaluated at budgets
$n\in\{50,100,200\}$. The reconstruction error obtained by Random at each
budget defines a target error. We then measure how many grasps FPS and
connectivity-aware sampling require to reach or improve upon the same
target. Fewer required grasps indicate greater structural reconstruction
efficiency.

\par\vspace{0.7em}
\paragraph*{Experiment 3: Hidden-Viability Exploration and Exploitation}

E3 evaluates whether structure inferred from partially observed grasp
outcomes improves the discovery of successful grasps in an unseen
candidate pool, and whether this effect depends on how the initial
observations are acquired. Candidate poses are known, but their
success/failure labels are revealed only when queried. We use two
disjoint phases: Phase~1 acquires observations from Pool~$\mathcal{A}$
using either random or connectivity-aware exploration, and Phase~2
compares alternative exploitation rules for ranking previously unqueried
candidates in Pool~$\mathcal{B}$.

For each object, we sample two disjoint balanced pools without
replacement from the ACRONYM grasp set. We evaluate
\[
N \in \{200,400,800,1600\},
\]
where $N=|\mathcal{A}|+|\mathcal{B}|$ and
$|\mathcal{A}|=|\mathcal{B}|=N/2$. Each pool contains equal numbers of
successful and failed grasps. Phase~1 and Phase~2 each query 50\% of
their respective pools, giving query budgets
\[
Q \in \{50,100,200,400\}.
\]
Object eligibility is fixed using the largest condition: each object
must contain at least 800 successful and 800 failed grasps, allowing the
same 200 objects to be used for all values of $N$. Each condition is
evaluated over five random seeds.

Distances use the normalized $SE(3)$ metric defined above.
Translational and rotational normalization scales are estimated from
Pool~$\mathcal{A}$ only and then fixed for all within- and cross-pool
distances, so Pool~$\mathcal{B}$ does not influence the geometric
representation used during exploration or exploitation.

\emph{Phase~1: observation acquisition.}
We consider two observation policies on Pool~$\mathcal{A}$:
\emph{Random} and \emph{connectivity-aware} (CA) exploration. Random
samples candidates uniformly without replacement. CA begins with five
random queries and subsequently selects candidates using the multiscale
bridge, frontier, boundary, and geometric-novelty criteria defined
above. Only successful queried grasps enter the viable connectivity
structure; failed queries consume the query budget but do not enter the
connectivity graph. Connectivity is evaluated over the same 15
neighborhood scales $\epsilon\in[0.10,0.80]$ used in E2. If no
successful grasp has yet been observed, CA continues random sampling.

For an exploration policy $e\in\{\mathrm{Random},\mathrm{CA}\}$, let
$S_{A,e}^{+}$ and $S_{A,e}^{-}$ denote the successful and failed grasps
observed during Phase~1. These observations are frozen before Phase~2.
Within each exploration condition, all exploitation methods therefore
receive exactly the same Phase~1 observations.

\emph{Phase~2: exploitation.}
For each Phase~1 observation policy, we compare four fixed exploitation
rules on the previously unseen Pool~$\mathcal{B}$: Random,
nearest-success (NS), local viability (LV), and connectivity-aware (CA).
Each rule produces a complete ranking of Pool~$\mathcal{B}$ before any
Phase~2 outcomes are observed, and the ranking is not updated during
Phase~2.

Random assigns a uniformly random ordering to Pool~$\mathcal{B}$ and
does not use the Phase~1 observations.

NS uses only the successful Phase~1 observations. For exploration policy
$e$, each candidate $x\in\mathcal{B}$ is scored by
\begin{equation}
d_{+,e}(x)
=
\min_{s\in S_{A,e}^{+}} d(x,s),
\end{equation}
and candidates are ranked in ascending order of $d_{+,e}(x)$.

LV uses both successful and failed Phase~1 observations. Defining
\begin{equation}
d_{+,e}(x)=\min_{s\in S_{A,e}^{+}}d(x,s),
\qquad
d_{-,e}(x)=\min_{f\in S_{A,e}^{-}}d(x,f),
\end{equation}
the local viability score is
\begin{equation}
S_{\mathrm{LV},e}(x)
=
\frac{d_{-,e}(x)-d_{+,e}(x)}
{d_{-,e}(x)+d_{+,e}(x)+\delta},
\qquad
\delta=10^{-12}.
\end{equation}
Candidates are ranked in descending order of
$S_{\mathrm{LV},e}(x)$, favoring candidates that are relatively closer
to observed successes than to observed failures.

CA uses the same successful and failed observations but additionally
accounts for the multiscale connectivity of the observed viable set.
At each scale $\epsilon\in\mathcal{E}$, the successful observations
$S_{A,e}^{+}$ form connected components
$\mathcal{C}_{\epsilon,e}$. Positive support for
$x\in\mathcal{B}$ is
\begin{equation}
\rho_{\epsilon,e}^{+}(x)
=
\max_{C\in\mathcal{C}_{\epsilon,e}}
\frac{1}{|S_{A,e}^{+}|}
\sum_{s\in C}
\exp\!\left(
-\frac{d(x,s)^2}{2\epsilon^2}
\right),
\end{equation}
while negative support is
\begin{equation}
\rho_{\epsilon,e}^{-}(x)
=
\frac{1}{|S_{A,e}^{-}|}
\sum_{f\in S_{A,e}^{-}}
\exp\!\left(
-\frac{d(x,f)^2}{2\epsilon^2}
\right).
\end{equation}
If no failed grasp is observed,
$\rho_{\epsilon,e}^{-}(x)=0$. The CA exploitation score is
\begin{equation}
S_{\mathrm{CA},e}(x)
=
\frac{1}{|\mathcal{E}|}
\sum_{\epsilon\in\mathcal{E}}
\frac{\rho_{\epsilon,e}^{+}(x)}
{\rho_{\epsilon,e}^{+}(x)+
 \rho_{\epsilon,e}^{-}(x)+\delta},
\qquad
\delta=10^{-12},
\end{equation}
where $\mathcal{E}$ contains the same 15 neighborhood scales
$\epsilon\in[0.10,0.80]$. Candidates are ranked in descending order of
$S_{\mathrm{CA},e}(x)$. If Phase~1 yields no successful grasp, methods
requiring successful observations fall back to a random ordering.

This design separates the effects of observation acquisition and
exploitation. Within each Phase~1 condition, NS, LV, and CA receive
identical observations, so their comparison isolates the exploitation
rule: NS tests success proximity alone, LV additionally incorporates
local failure evidence, and CA further incorporates multiscale
connectivity. Comparing the same exploitation rule after Random and CA
exploration tests whether connectivity-aware acquisition produces
observations that are more useful for subsequent grasp discovery.

The primary outcome is Phase~2 discovery AUC, computed from the
cumulative number of successful grasps versus query count, including the
origin. Phase~1 outcomes are used only to construct the Phase~2 rankings
and do not contribute directly to this metric.
\par\vspace{0.7em}

\paragraph*{Experiment 4: Connectivity-Aware Cross-Object Grasp Exploration}

E4 investigates whether connectivity-informed reference memories and
structural retrieval improve sequential grasp exploration on unseen
objects. We examine two mechanisms: selecting prototypes from
connected regions of successful reference grasps and retrieving them
using local structural signatures rather than absolute pose proximity.

\textit{Experimental protocol.}
Eligible objects are partitioned into disjoint reference, validation,
and evaluation pools. Reference objects also provide supervised
training data, $\mathcal D_{\mathrm{train}}=\mathcal D_{\mathrm{ref}}$.
A fixed partition seed defines the pools; independent runs construct
memories and train encoders with different seeds. Evaluation objects
may overlap across runs. No category-level separation is imposed;
E4 therefore evaluates transfer to held-out objects, not unseen
categories. Pose preprocessing and the weighted Euclidean metric
$d_\rho$ are defined in Appendix~\ref{app:e4_pose}.

\textit{Reference memory and structural representation.}
For each reference object, successful grasps form an
$\varepsilon$-neighborhood graph under $d_\rho$. Components with
fewer than two grasps are discarded, and medoid-initialized
farthest-point sampling selects up to $P$ prototypes per object.
The resulting memory is
\begin{equation}
\mathcal M=\{(p_i,o_i,c_i)\}_{i=1}^{K},
\qquad K\leq N_{\mathrm{ref}}P,
\label{eq:e4_main_memory}
\end{equation}
where $p_i$, $o_i$, and $c_i$ denote prototype pose, source object,
and component identity, respectively.

Each candidate $x$ also receives an eight-dimensional structural
signature $s(x;X)$ computed from the full unlabeled candidate-pose
graph of its object. These signatures describe local neighborhood
geometry, not the unknown connectivity of successful target grasps.
Reference prototypes inherit signatures from their source objects.
Memory construction and signature definitions are detailed in
Appendices~\ref{app:e4_pose} and~\ref{app:e4_signatures}.

\textit{Methods and mechanism comparisons.}
We compare six policies:
\emph{Random}, uniform sampling of untested candidates;
\emph{Target CA}, exploration using target observations without
reference memory;
\emph{Random geometry}, random successful reference prototypes
with pose-space retrieval;
\emph{FPS geometry}, geometrically dispersed prototypes with
pose-space retrieval;
\emph{Geometry encoder}, connectivity-selected prototypes with
pose-space retrieval; and
\emph{Connectivity encoder}, the same prototypes with
structural-signature retrieval.

The four learned policies share an architecture and training
protocol, with count-matched per-object reference memories.
These comparisons assess geometric dispersion (FPS versus random),
connectivity-informed selection (geometry versus FPS), and
structural retrieval (connectivity versus geometry). The final
comparison changes the retrieval representation and distance,
rather than isolating individual signature channels.

\textit{Outcome-conditioned exploration.}
Let $H_t=(S_t^+,S_t^-)$ contain successful and failed target grasps
observed before step $t$. The connectivity encoder dynamically
reranks structural neighbors using Gaussian evidence from $H_t$.
In the dynamic-geometric configurations evaluated here, the three
geometric encoders analogously rerank pose-space neighbors.
Only previously observed target outcomes are used; reference
memories and encoder parameters remain fixed.

Each learned policy predicts
\begin{equation}
\widehat p_{\theta}(x\mid H_t,\mathcal M)
=\sigma\!\left(g_{\theta}\!\left[
\phi(x),h_t(x),z_{\mathcal M}(x,H_t)
\right]\right),
\label{eq:e4_main_predictor}
\end{equation}
where $\phi(x)$ represents the candidate pose, $h_t(x)$ summarizes
target observations, and $z_{\mathcal M}$ encodes retrieved reference
features through permutation-invariant pooling.

Encoders are trained with binary cross-entropy on sampled partial
observation histories, leave-one-object-out reference masking, and
held-out validation. Training histories are not on-policy rollouts.
At evaluation, the highest-scoring untested grasp is selected:
\begin{equation}
x_t^*=\arg\max_{x\in U_t}
\widehat p_{\theta}(x\mid H_t,\mathcal M).
\label{eq:e4_main_selection}
\end{equation}
Its outcome updates $H_t$ and the applicable retrieval rankings,
without gradient updates. 
Architectural and retrieval details are provided in
Appendix~\ref{app:e4_structural_retrieval},
while training and evaluation procedures are detailed in
Appendix~\ref{app:e4_algorithm} and
Algorithms~\ref{alg:e4_training}
and~\ref{alg:e4_evaluation}.

\textit{Metrics and scaling.}
Each policy performs at most $B=100$ grasp attempts per object.
We measure the number of successes and the mean cumulative-success
area:
\begin{equation}
C_B=\sum_{t=1}^{B}y_t,
\qquad
\mathrm{AUC}_{\mathrm{succ}}
=\frac{1}{B}\sum_{t=1}^{B}\sum_{\tau=1}^{t}y_\tau,
\label{eq:e4_main_metrics}
\end{equation}
where $y_t\in\{0,1\}$ is the observed outcome. AUC rewards earlier
successes; when fewer candidates are available, the realized query
count replaces $B$.

We conduct two scaling studies. First, reference and training
objects jointly vary over
$N_{\mathrm{ref}}=N_{\mathrm{train}}\in
\{6,12,25,50,100,200,400,800\}$ with $P=16$.
This changes training size, reference diversity, and memory capacity
simultaneously. Second, we fix $N_{\mathrm{ref}}=50$ and vary
$P\in\{4,8,16,32,64\}$, isolating the effect of nominal
per-object memory capacity. Actual prototype counts are recorded.

Each configuration uses five independently trained seeds and
500 evaluation objects per seed. Methods share evaluation targets
and candidate sets within each seed. We aggregate outcomes first
over targets, then report means and sample standard deviations
across seeds. Statistical comparisons use paired seed-level
differences, avoiding pseudoreplication from targets sharing
the same trained encoder.

\section{Results}

\subsection{E1: Reproducibility and Object Specificity}
\label{sec:e1_results}

\begin{figure}[t]
    \centering
    \includegraphics[width=\columnwidth]{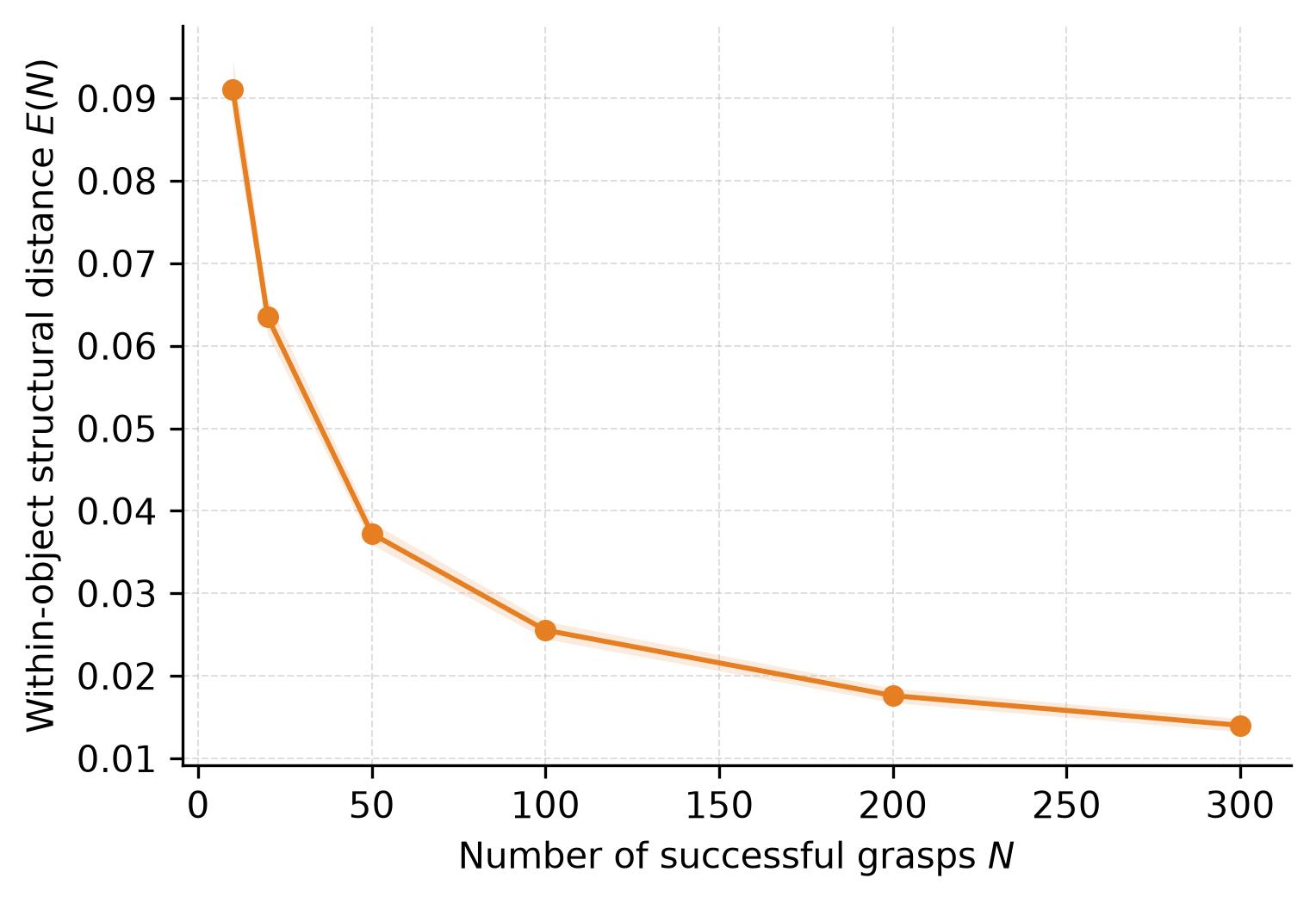}
    \caption{
    Convergence of the grasp-space connectivity signature with sample size.
    The curve shows the mean within-object structural distance $E(N)$
    across 100 objects, with 95\% bootstrap confidence intervals across
    objects. Independent samples become increasingly consistent as the
    number of successful grasps increases.
    }
    \label{fig:e1_convergence}
\end{figure}

\begin{figure}[t]
    \centering
    \includegraphics[width=\columnwidth]{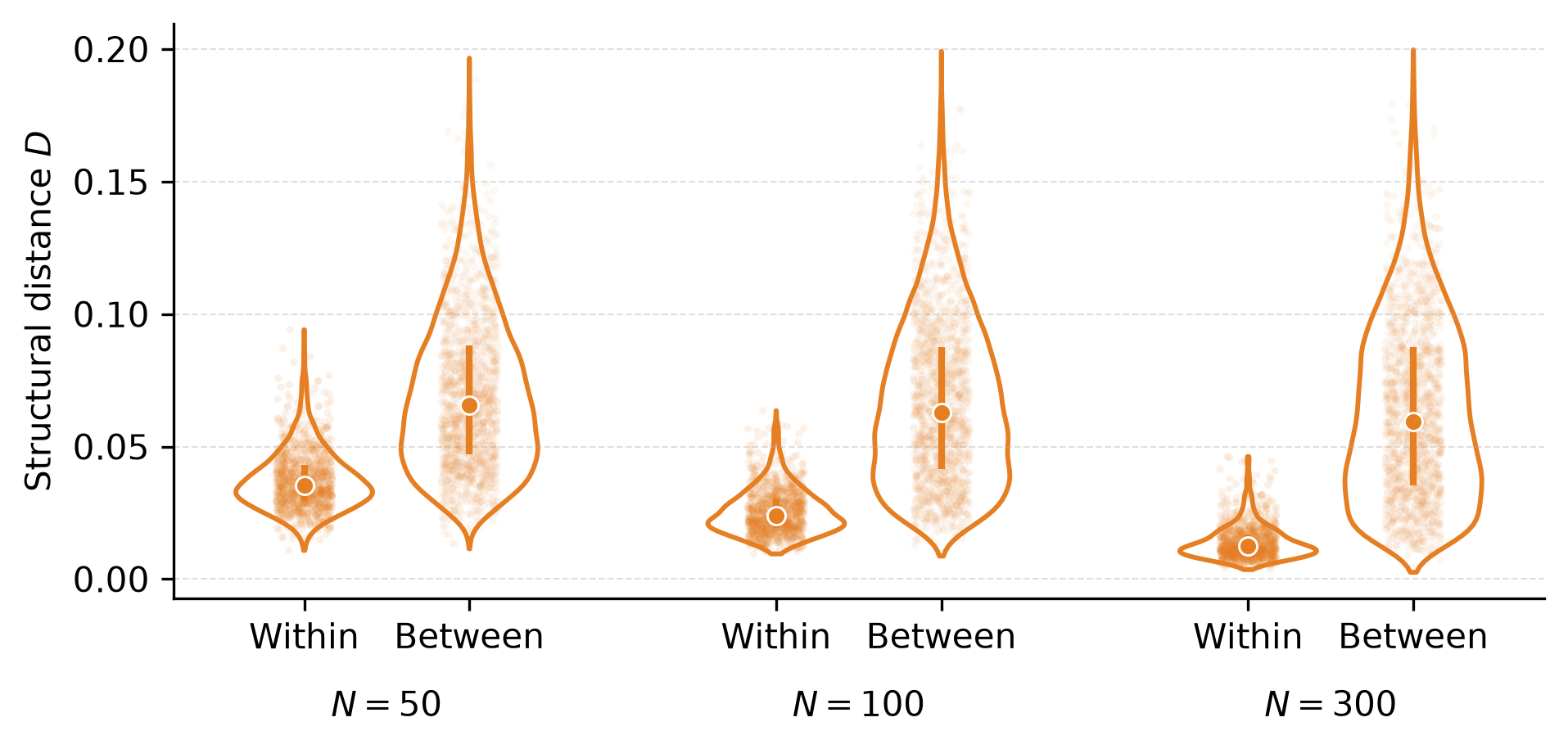}
    \caption{
    Within-object and between-object structural distances at
    $N=50$, $100$, and $300$ successful grasps.
    Within-object distances decrease strongly with sample size, whereas
    between-object distances remain substantially larger and comparatively
    stable. Sampling variability therefore decreases with $N$ while
    object-specific differences in multiscale connectivity persist.
    Vertical lines show the interquartile range and circles denote medians.
    }
    \label{fig:e1_within_between}
\end{figure}
Experiment 1 (E1) tested whether multiscale connectivity is a
reproducible and object-specific property of viable grasp spaces. The
results support both properties: connectivity signatures became
progressively more consistent across independent samples of the same
object as sample size increased, while differences between objects
persisted.

\begin{figure*}[t]
    \centering

    \includegraphics[width=0.325\textwidth]{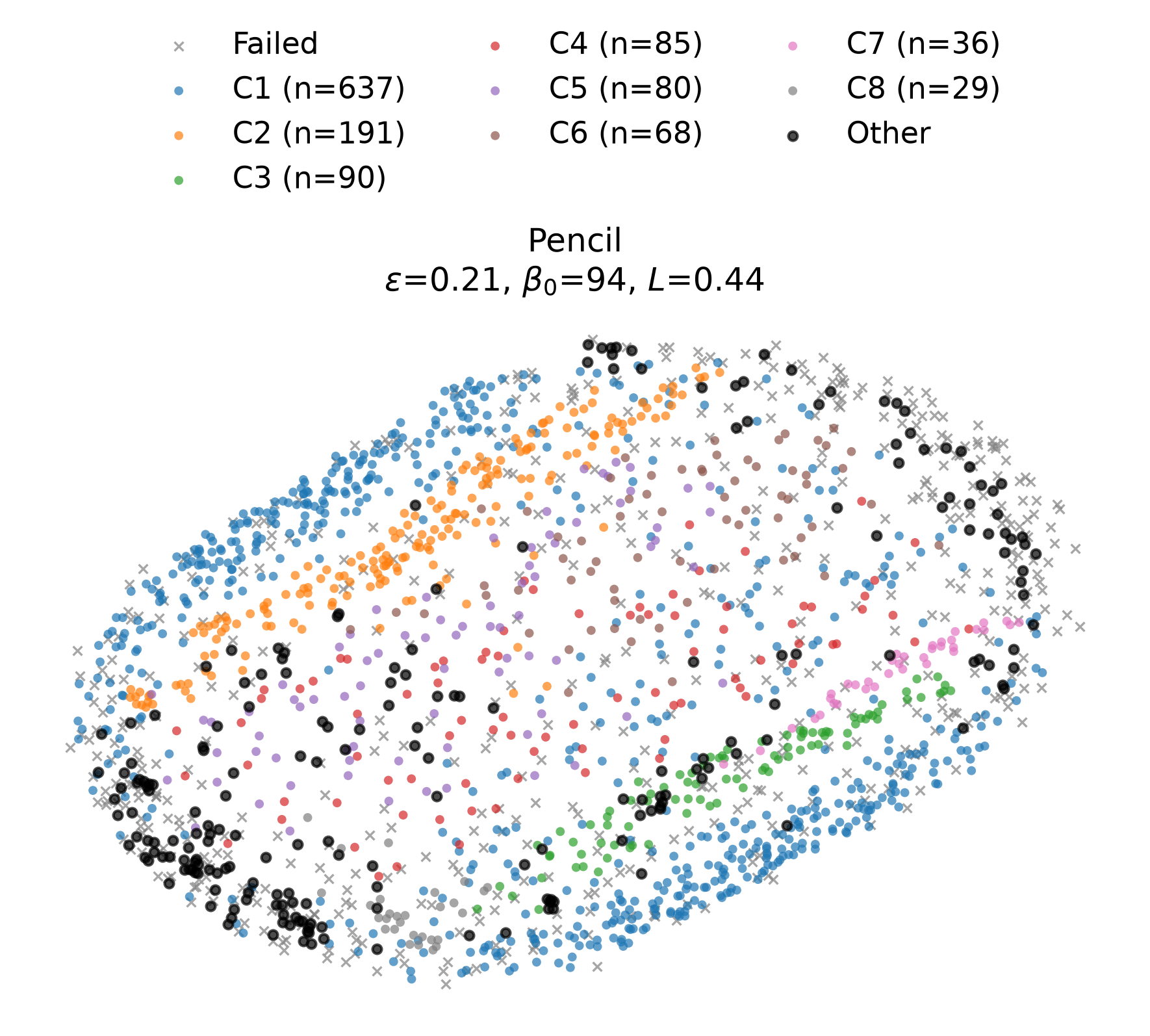}
    \hfill
    \includegraphics[width=0.325\textwidth]{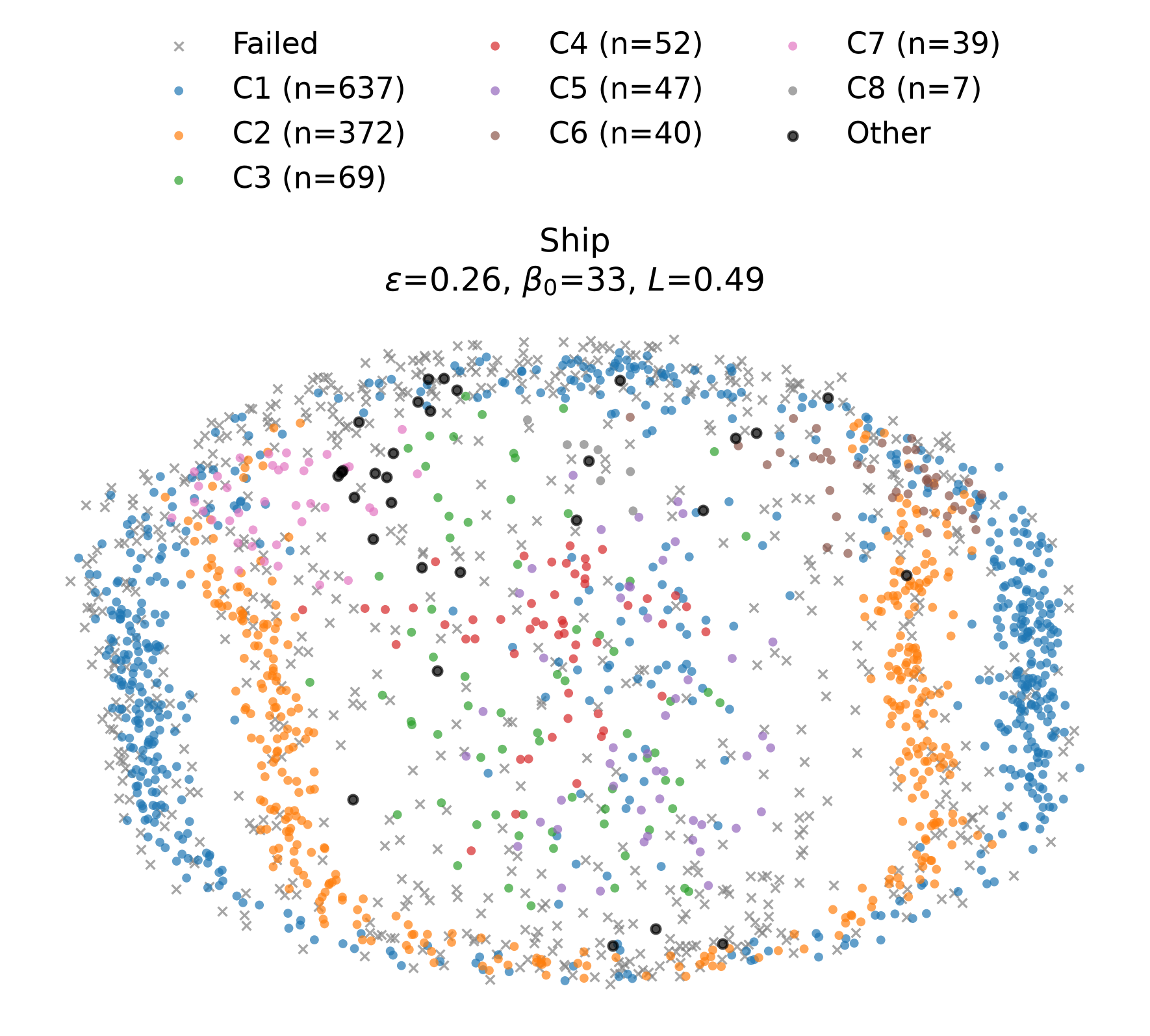}
    \hfill
    \includegraphics[width=0.325\textwidth]{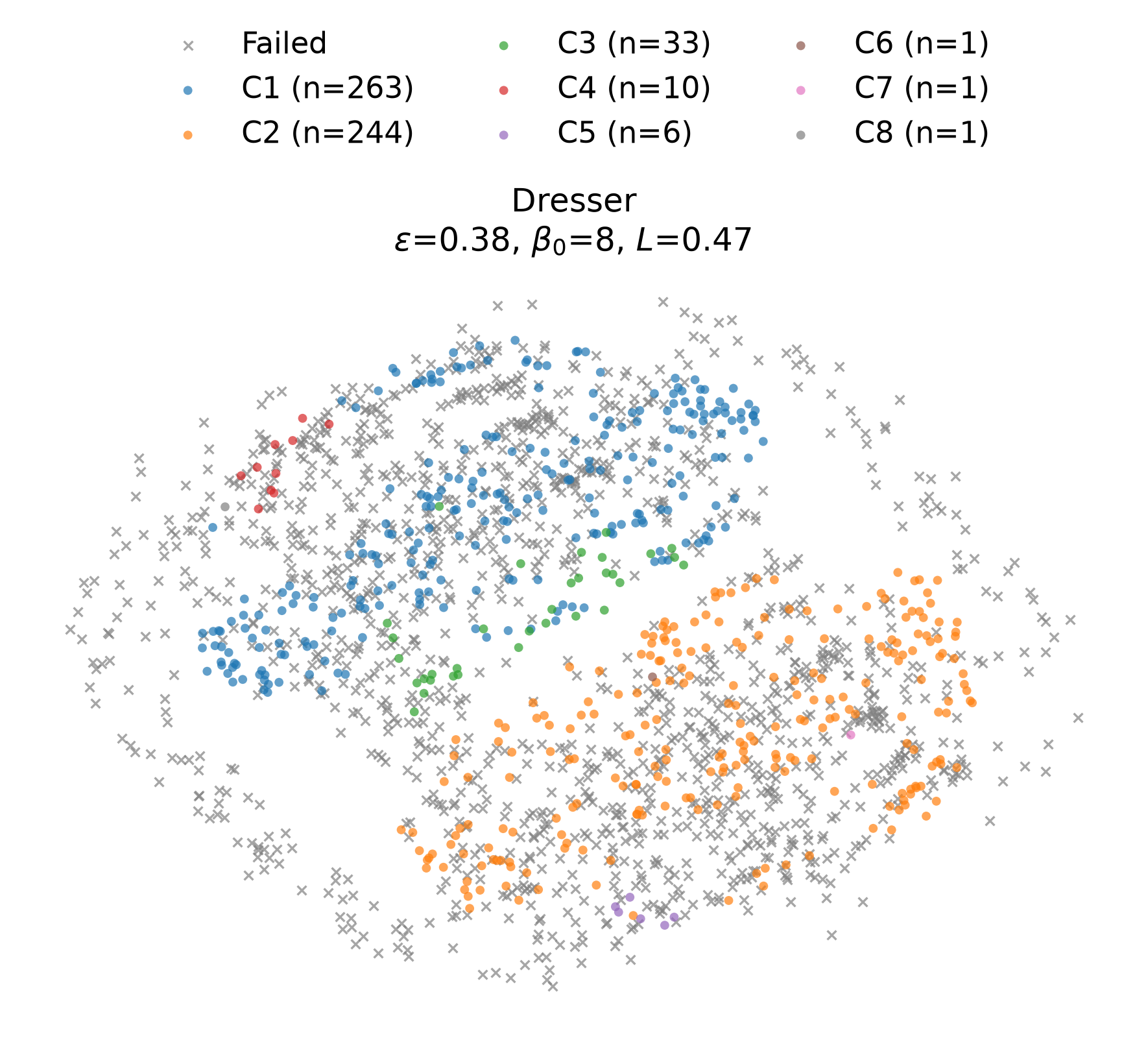}

    \vspace{0.4em}

    \includegraphics[width=0.325\textwidth]{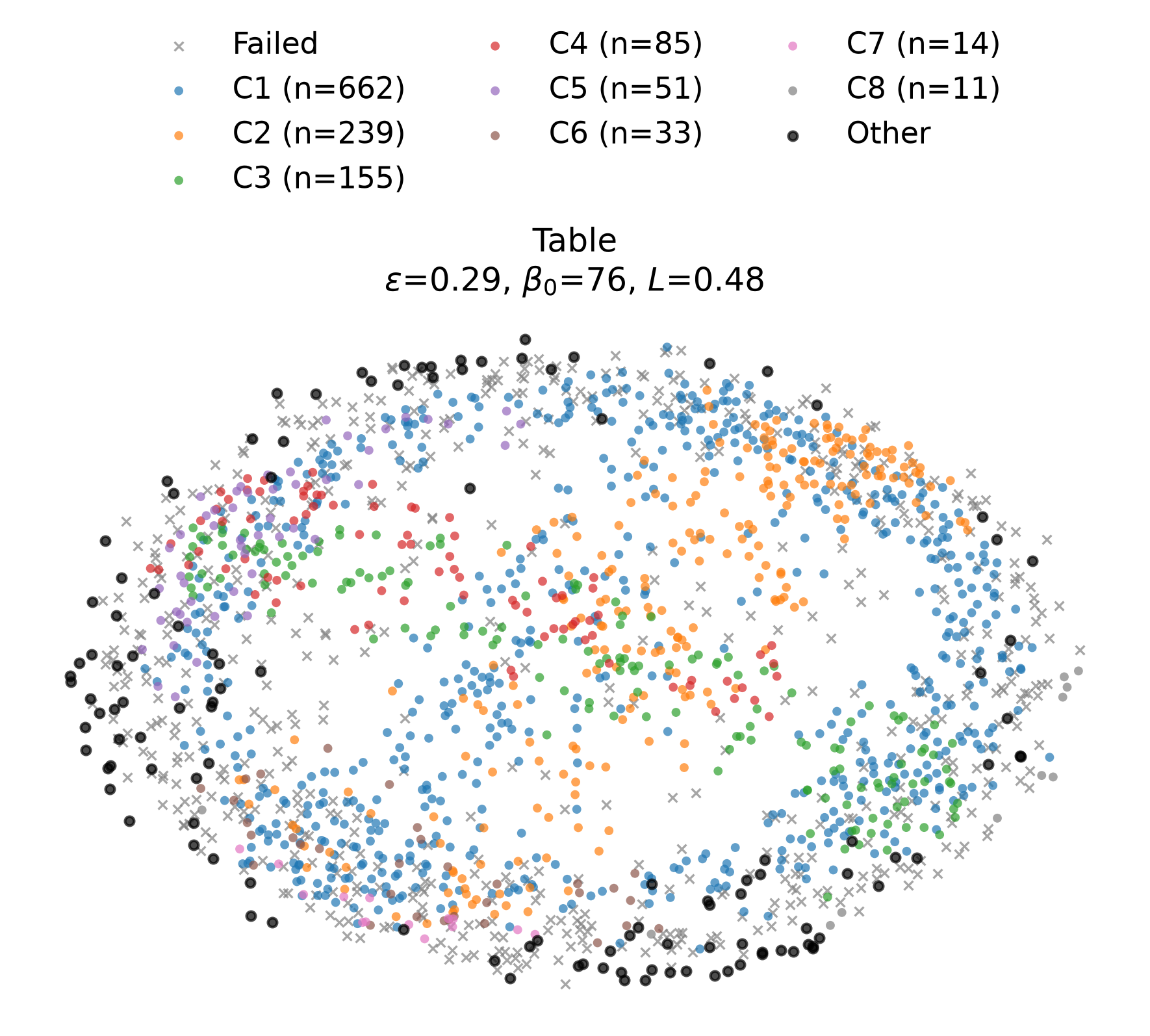}
    \hfill
    \includegraphics[width=0.325\textwidth]{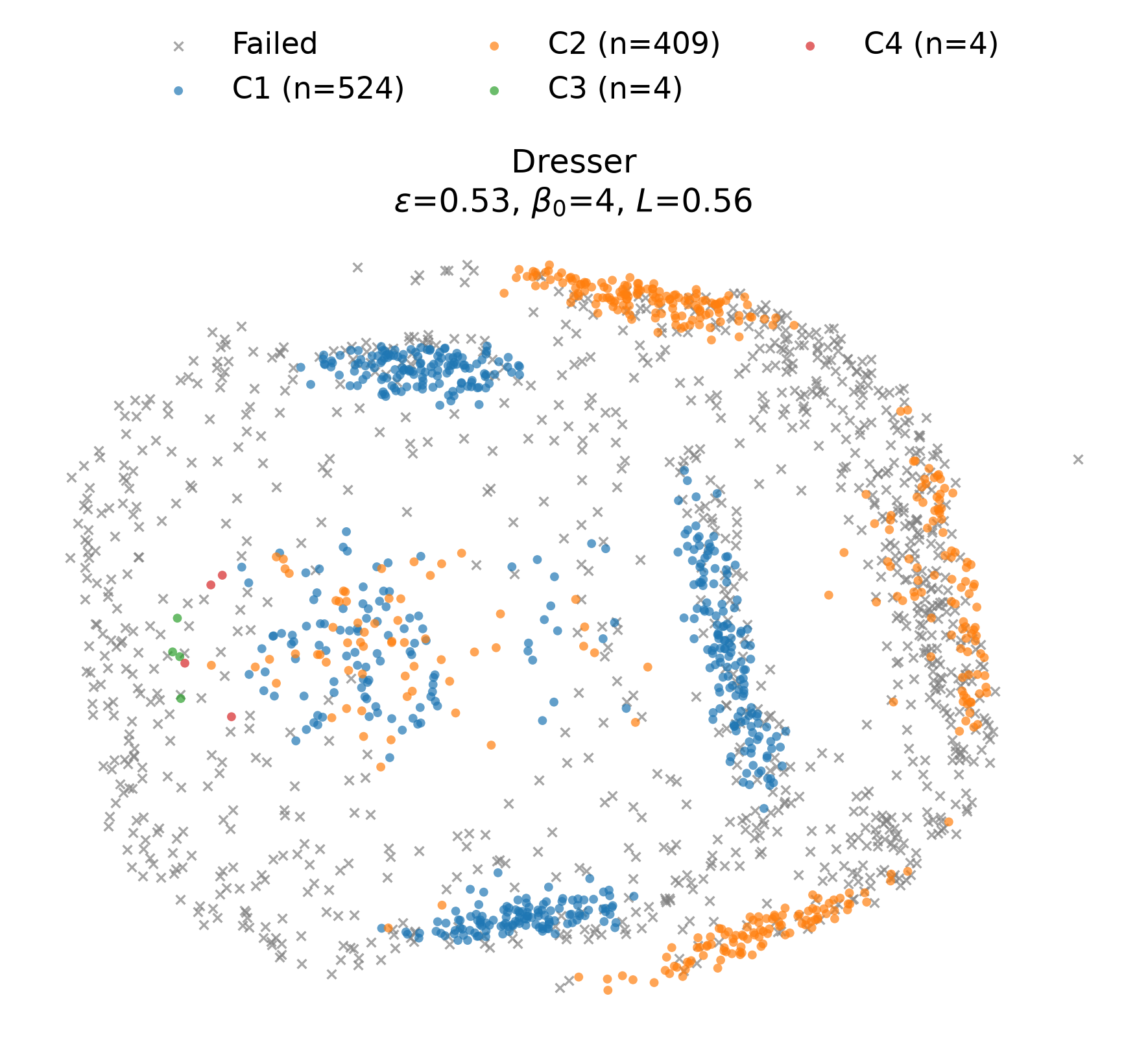}
    \hfill
    \includegraphics[width=0.325\textwidth]{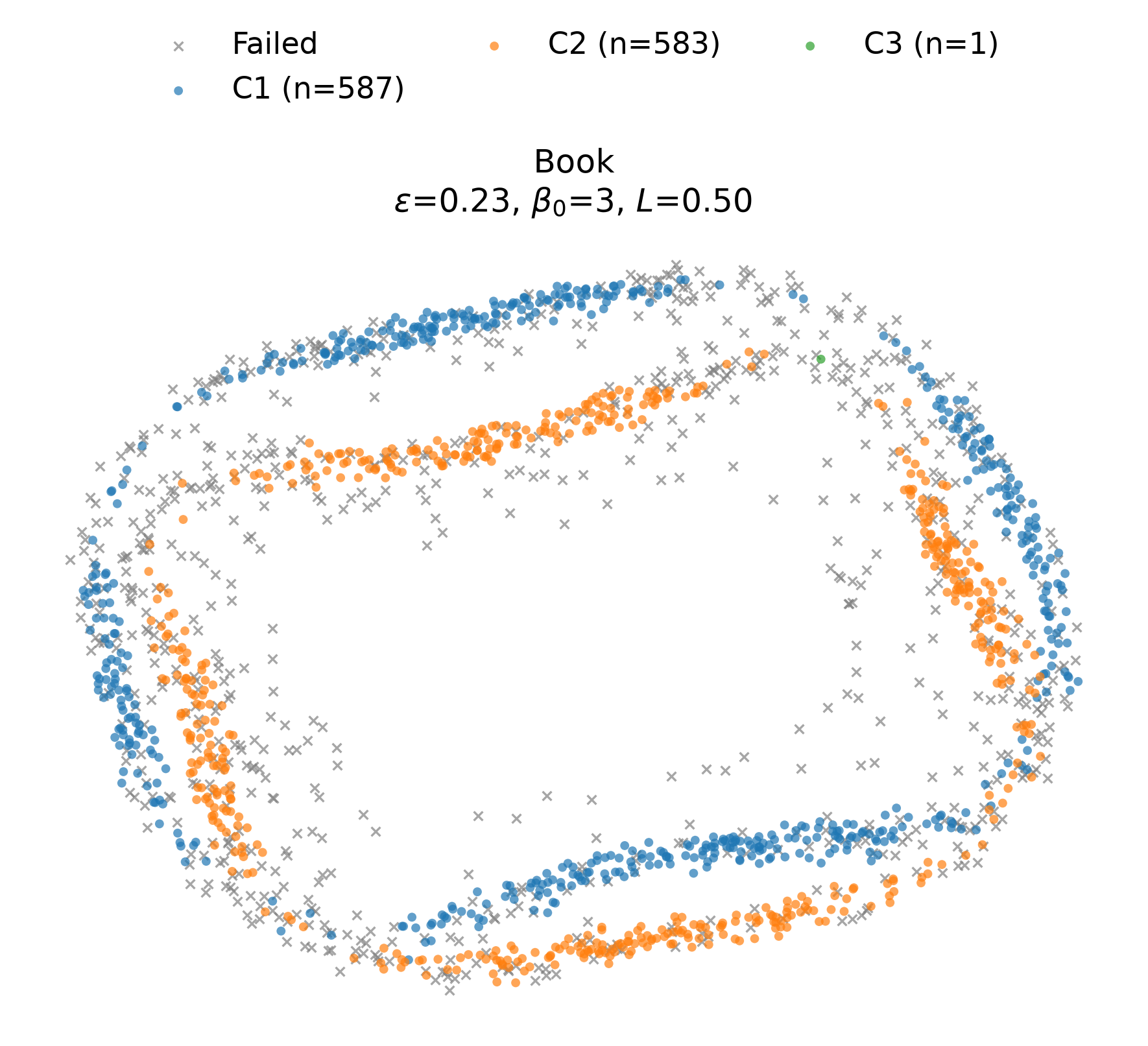}

    \caption{
    Examples of viable grasp-space structure across six objects.
    Successful grasps are colored according to connected components
   computed in the original normalized $SE(3)$ grasp space, while failed
    grasps are shown in gray. The two-dimensional MDS embedding is used
    only for visualization. Each panel reports the connectivity radius
    $\epsilon$, number of connected components $\beta_0$, and relative
    size $L$ of the largest component.
    }
    \label{fig:grasp_space_examples}
\end{figure*}
 As shown in
Fig.~\ref{fig:e1_convergence}, the mean structural distance between
independent samples of the same object decreased monotonically as the
number of successful grasps increased. The mean within-object distance
was $0.0372$ at $N=50$, $0.0255$ at $N=100$, $0.0176$ at $N=200$, and
$0.0140$ at $N=300$, with corresponding 95\% bootstrap confidence
intervals of $[0.0359,0.0386]$, $[0.0245,0.0266]$,
$[0.0167,0.0185]$, and $[0.0132,0.0148]$. The same decreasing trend
extends across the complete range of sample sizes tested, from $N=10$
to $N=300$. Independent samples of an object's viable grasp set
therefore converge toward increasingly similar multiscale connectivity
signatures as more successful grasps are observed.

Importantly, the reduction in sampling variability did not eliminate
differences between objects. Figure~\ref{fig:e1_within_between} compares
distances between independent samples from the same object with distances
between samples from different objects at $N=50$, $100$, and $300$.
Within-object distances contract strongly as $N$ increases, consistent
with the convergence analysis. In contrast, the between-object
distribution remains comparatively stable and substantially above the
within-object distribution. Consequently, the separation between
within-object and between-object structure becomes increasingly
pronounced as sample size grows.

\begin{table*}[t]
\centering
\caption{Hidden-viability exploitation under Random and connectivity-aware
(CA) Phase~1 exploration across candidate-set sizes. Phase~2 success counts
and normalized AUC are reported as mean $\pm$ SD across objects after
averaging over five seeds. Gains are computed relative to Random exploitation
under the same Phase~1 exploration policy and reported as mean [95\% CI].
NS denotes nearest-success and LV local viability.}
\label{tab:e3_multiscale}

\begin{tabular}{cclccc}
\hline
$N$ & Exploration & Method
& Phase~2 successes
& Normalized AUC
& Mean gain [95\% CI] (\%) \\
\hline

200 & Random & Random
& $25.13 \pm 1.14$
& $0.5020 \pm 0.0295$
& -- \\

& & NS
& $30.97 \pm 4.45$
& $0.6638 \pm 0.1054$
& $+32.66\;[+29.62,\,+35.80]$ \\

& & LV
& {\boldmath$31.20 \pm 4.16$}
& {\boldmath$0.6699 \pm 0.1007$}
& {\boldmath$+33.94\;[+30.93,\,+37.04]$} \\

& & CA
& $31.02 \pm 4.18$
& $0.6528 \pm 0.0983$
& $+30.56\;[+27.61,\,+33.57]$ \\

\cline{2-6}

& CA & Random
& $25.13 \pm 1.14$
& $0.5020 \pm 0.0295$
& -- \\

& & NS
& $29.77 \pm 4.47$
& $0.6526 \pm 0.1050$
& $+30.44\;[+27.39,\,+33.51]$ \\

& & LV
& {\boldmath$30.56 \pm 4.23$}
& {\boldmath$0.6627 \pm 0.1001$}
& {\boldmath$+32.49\;[+29.52,\,+35.48]$} \\

& & CA
& $30.32 \pm 4.20$
& $0.6434 \pm 0.0989$
& $+28.66\;[+25.71,\,+31.59]$ \\

\hline

400 & Random & Random
& $49.75 \pm 1.61$
& $0.4981 \pm 0.0204$
& -- \\

& & NS
& $64.25 \pm 8.23$
& $0.6822 \pm 0.0937$
& $+37.21\;[+34.46,\,+40.02]$ \\

& & LV
& $65.06 \pm 7.76$
& {\boldmath$0.7037 \pm 0.0905$}
& {\boldmath$+41.47\;[+38.90,\,+44.18]$} \\

& & CA
& {\boldmath$65.14 \pm 7.69$}
& $0.6928 \pm 0.0903$
& $+39.26\;[+36.72,\,+41.85]$ \\

\cline{2-6}

& CA & Random
& $49.75 \pm 1.61$
& $0.4981 \pm 0.0204$
& -- \\

& & NS
& $62.86 \pm 8.41$
& $0.6760 \pm 0.0937$
& $+36.01\;[+33.22,\,+38.82]$ \\

& & LV
& $63.86 \pm 7.33$
& {\boldmath$0.6960 \pm 0.0893$}
& {\boldmath$+39.95\;[+37.37,\,+42.54]$} \\

& & CA
& {\boldmath$64.04 \pm 7.52$}
& $0.6836 \pm 0.0898$
& $+37.40\;[+34.88,\,+39.95]$ \\

\hline

800 & Random & Random
& $100.12 \pm 2.36$
& $0.5004 \pm 0.0159$
& -- \\

& & NS
& $132.03 \pm 16.33$
& $0.7021 \pm 0.0900$
& $+40.41\;[+37.92,\,+42.97]$ \\

& & LV
& $135.06 \pm 16.19$
& {\boldmath$0.7315 \pm 0.0927$}
& {\boldmath$+46.26\;[+43.67,\,+48.81]$} \\

& & CA
& {\boldmath$135.74 \pm 15.60$}
& $0.7254 \pm 0.0892$
& $+45.04\;[+42.58,\,+47.54]$ \\

\cline{2-6}

& CA & Random
& $100.12 \pm 2.36$
& $0.5004 \pm 0.0159$
& -- \\

& & NS
& $130.77 \pm 15.55$
& $0.6941 \pm 0.0864$
& $+38.78\;[+36.40,\,+41.20]$ \\

& & LV
& $135.16 \pm 15.64$
& {\boldmath$0.7314 \pm 0.0896$}
& {\boldmath$+46.24\;[+43.78,\,+48.72]$} \\

& & CA
& {\boldmath$135.72 \pm 15.48$}
& $0.7221 \pm 0.0898$
& $+44.37\;[+41.90,\,+46.87]$ \\

\hline

1600 & Random & Random
& $199.61 \pm 3.30$
& $0.4995 \pm 0.0109$
& -- \\

& & NS
& $269.79 \pm 33.04$
& $0.7187 \pm 0.0892$
& $+43.96\;[+41.45,\,+46.51]$ \\

& & LV
& $279.53 \pm 31.59$
& {\boldmath$0.7614 \pm 0.0887$}
& {\boldmath$+52.52\;[+50.09,\,+55.06]$} \\

& & CA
& {\boldmath$281.10 \pm 29.78$}
& $0.7598 \pm 0.0845$
& $+52.19\;[+49.83,\,+54.53]$ \\

\cline{2-6}

& CA & Random
& $199.61 \pm 3.30$
& $0.4995 \pm 0.0109$
& -- \\

& & NS
& $264.67 \pm 30.93$
& $0.7005 \pm 0.0827$
& $+40.32\;[+37.99,\,+42.68]$ \\

& & LV
& $279.27 \pm 30.66$
& $0.7563 \pm 0.0857$
& $+51.49\;[+49.08,\,+53.94]$ \\

& & CA
& {\boldmath$282.66 \pm 29.43$}
& {\boldmath$0.7577 \pm 0.0823$}
& {\boldmath$+51.77\;[+49.47,\,+54.09]$} \\

\hline
\end{tabular}

\vspace{0.3em}
\footnotesize
$N$ denotes the total number of candidates across the disjoint exploration
and exploitation pools; the query budget in each phase is $K=N/4$.
Within each exploration condition, all Phase~2 methods receive identical
Phase~1 observations. Results are first averaged over five seeds within each
object and then summarized across 200 objects. Standard deviations quantify
object-to-object variation. Gain confidence intervals are 95\% paired
bootstrap intervals across objects. Bold indicates the best value separately
within each Random- and CA-exploration block.
\end{table*}
Fig.~\ref{fig:grasp_space_examples} illustrates the heterogeneity of
viable grasp-space organization across representative objects. Connected
components were computed in the original normalized $SE(3)$ grasp space;
for visualization only, the corresponding pairwise distance matrix was
embedded in two dimensions using metric multidimensional scaling
(MDS)~\cite{borg2005modern}. For each object, the visualization radius
$\epsilon$ was selected so that the largest connected component contained
approximately half of the successful grasps. This constrains only the size
of the largest component, not the number or sizes of the remaining
components. Consequently, objects with similar largest-component occupancy
can still exhibit substantially different connectivity structure, ranging
from a few large connected regions to many smaller components.

\subsection{Exploration Efficiency}
\label{sec:exploration_efficiency}

Connectivity-aware sampling reduced the number of viable grasp observations
required to recover the reference connectivity structure
(Table~\ref{tab:e2_reconstruction}). At the Random@50 and Random@100
reconstruction-error targets, CA reached the corresponding error level for
all 200 objects, requiring median sample counts of 12 and 21 observations,
respectively. These correspond to median object-level speedups of
$4.17\times$ and $4.76\times$. At the more stringent Random@200 target,
CA reached the corresponding error level within the 200-observation budget
for 192 of 200 objects (96\%), requiring a median of 93 observations among
the objects that reached the target, corresponding to a conditional median
speedup of $2.15\times$.

\begin{table}[t]
\centering
\caption{Sample efficiency of connectivity-aware (CA) exploration in E2.
For each Random sampling budget, we report the median number of CA
observations required to achieve the same connectivity-reconstruction
error. Speedup is computed per object as the Random budget divided by
the required CA observations and then summarized by the median.}
\label{tab:e2_reconstruction}
\small
\setlength{\tabcolsep}{4pt}
\begin{tabular}{lccc}
\hline
\shortstack{Random\\observations} &
\shortstack{Objects matched\\by CA} &
\shortstack{Median CA\\observations} &
\shortstack{Median\\speedup} \\
\hline
50  & 200/200 & 12 & $4.17\times$ \\
100 & 200/200 & 21 & $4.76\times$ \\
200 & 192/200 & 93 & $2.15\times$ \\
\hline
\end{tabular}
\end{table}

\begin{figure*}[!t]
    \centering

    \begin{subfigure}[t]{0.48\textwidth}
        \centering
        \includegraphics[width=\linewidth]{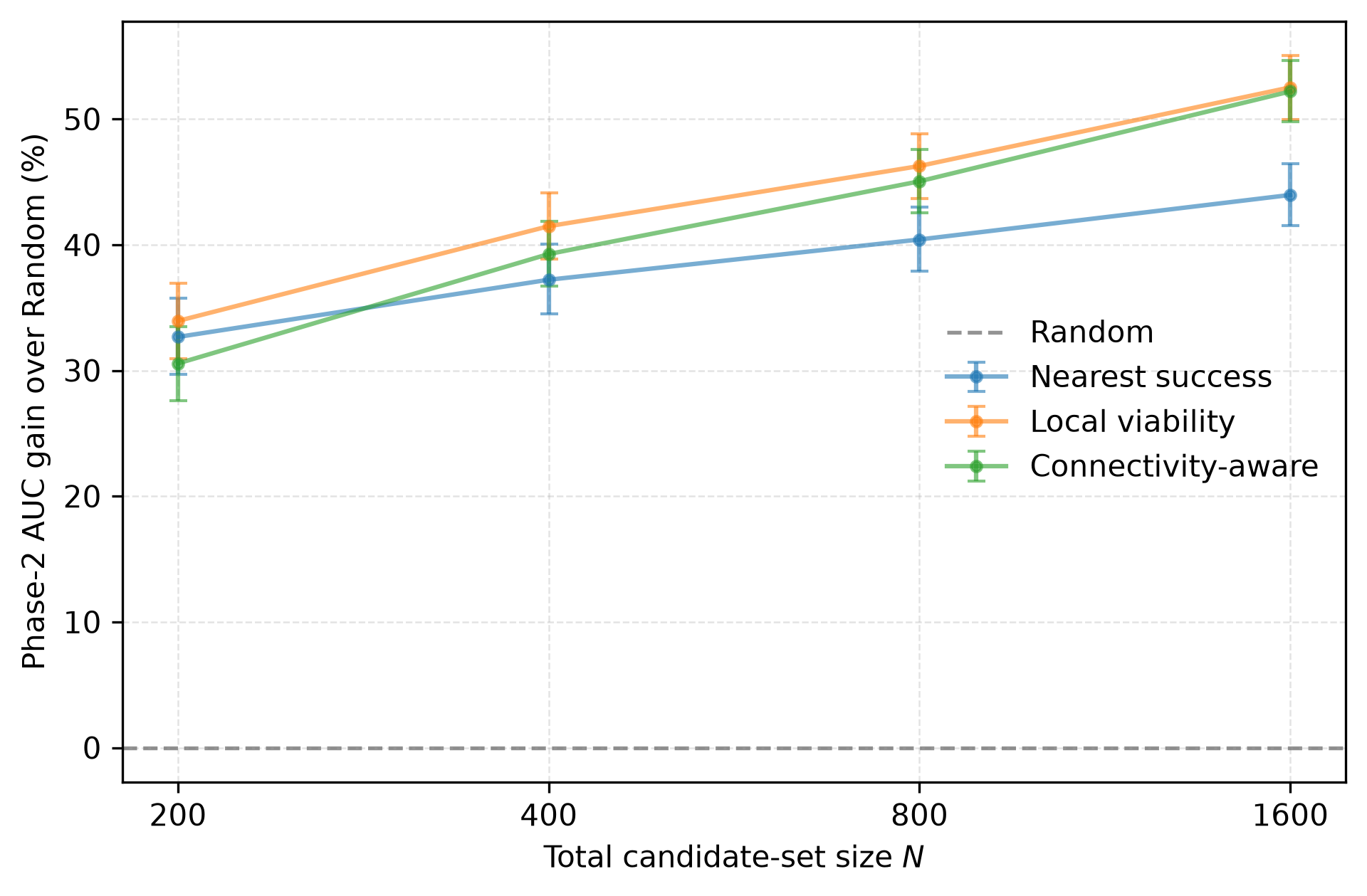}
        \caption{Scaling with candidate-set size.}
        \label{fig:e3_ca_scaling}
    \end{subfigure}
    \hfill
    \begin{subfigure}[t]{0.48\textwidth}
        \centering
        \includegraphics[width=\linewidth]{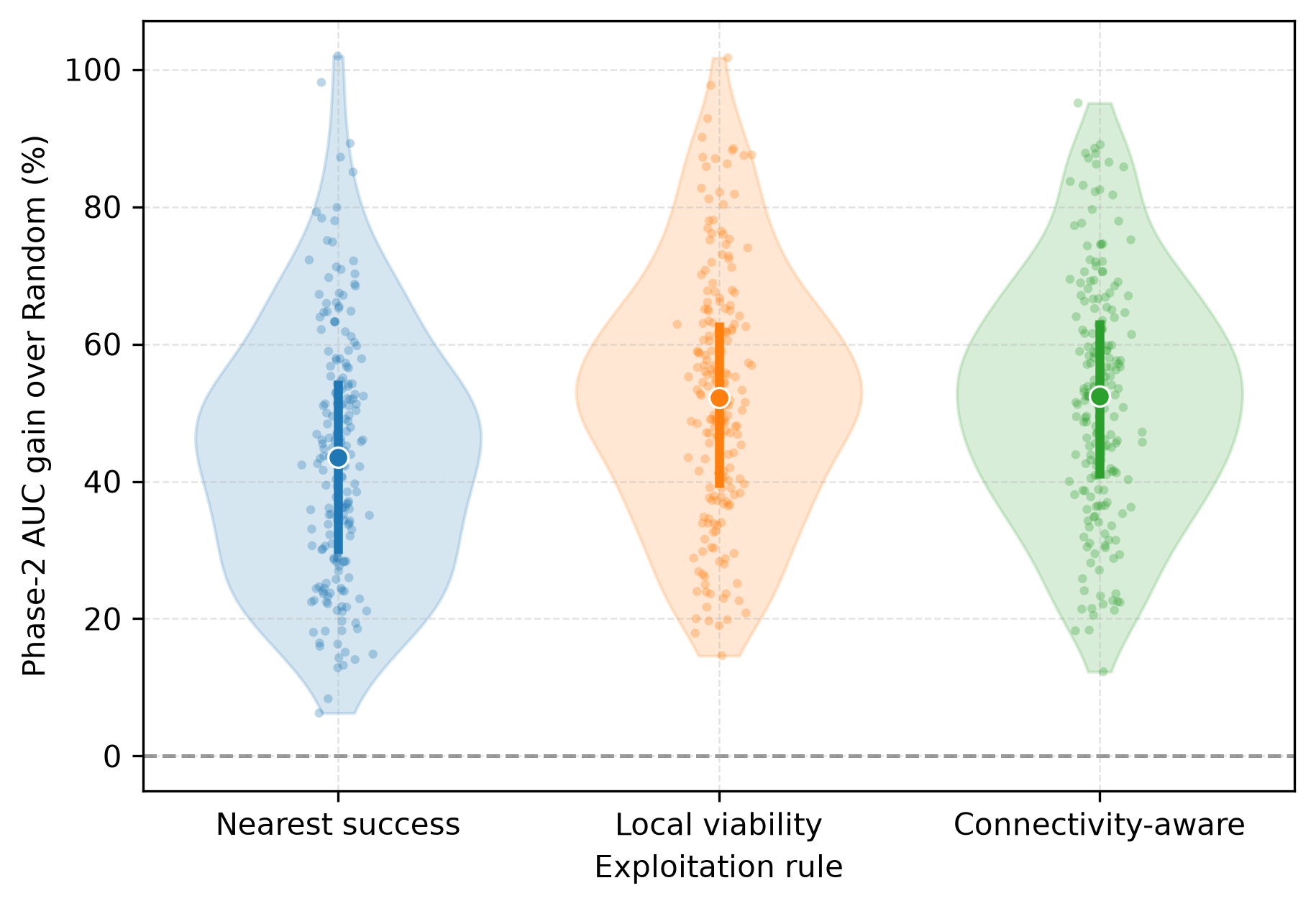}
        \caption{Object-level gains at $N=1600$.}
        \label{fig:e3_ca_distribution}
    \end{subfigure}

    \caption{
    Phase~2 exploitation performance following connectivity-aware (CA)
    Phase~1 exploration. Left: mean Phase~2 AUC gain relative to Random
    exploitation as a function of total candidate-set size $N$; error bars
    show 95\% bootstrap confidence intervals across objects. Right:
    distributions of object-level Phase~2 AUC gains relative to Random
    exploitation at $N=1600$; points denote individual objects, circles
    indicate medians, and vertical bars indicate interquartile ranges.
    }
    \label{fig:e3_ca_exploitation}
\end{figure*}

\begin{figure*}[t]
    \centering

    \begin{subfigure}[t]{0.325\textwidth}
        \centering
        \includegraphics[width=\linewidth]{
            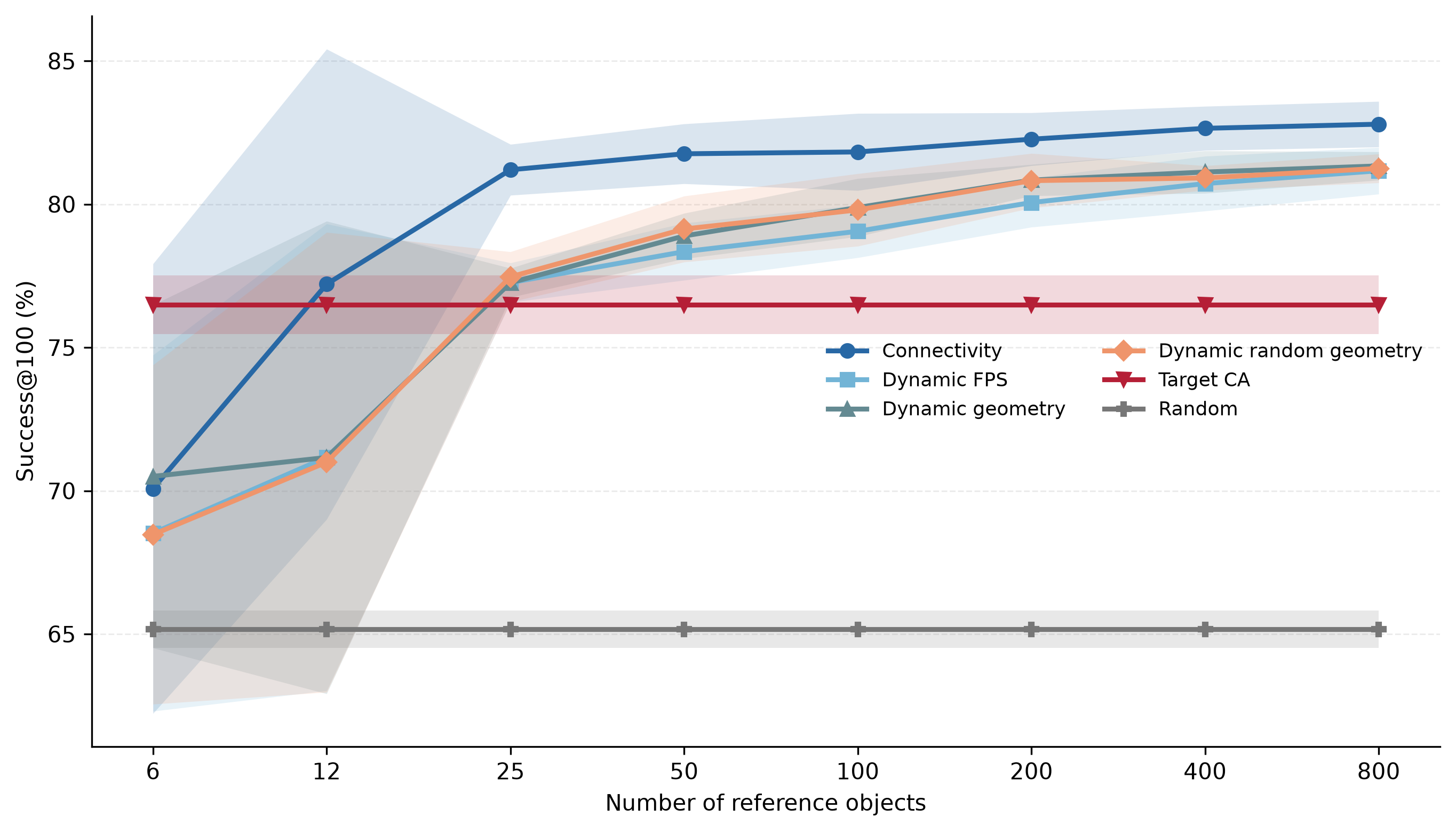
        }
        \caption{Reference-object scaling.}
        \label{fig:e4_reference_scaling}
    \end{subfigure}
    \hfill
    \begin{subfigure}[t]{0.325\textwidth}
        \centering
        \includegraphics[width=\linewidth]{
            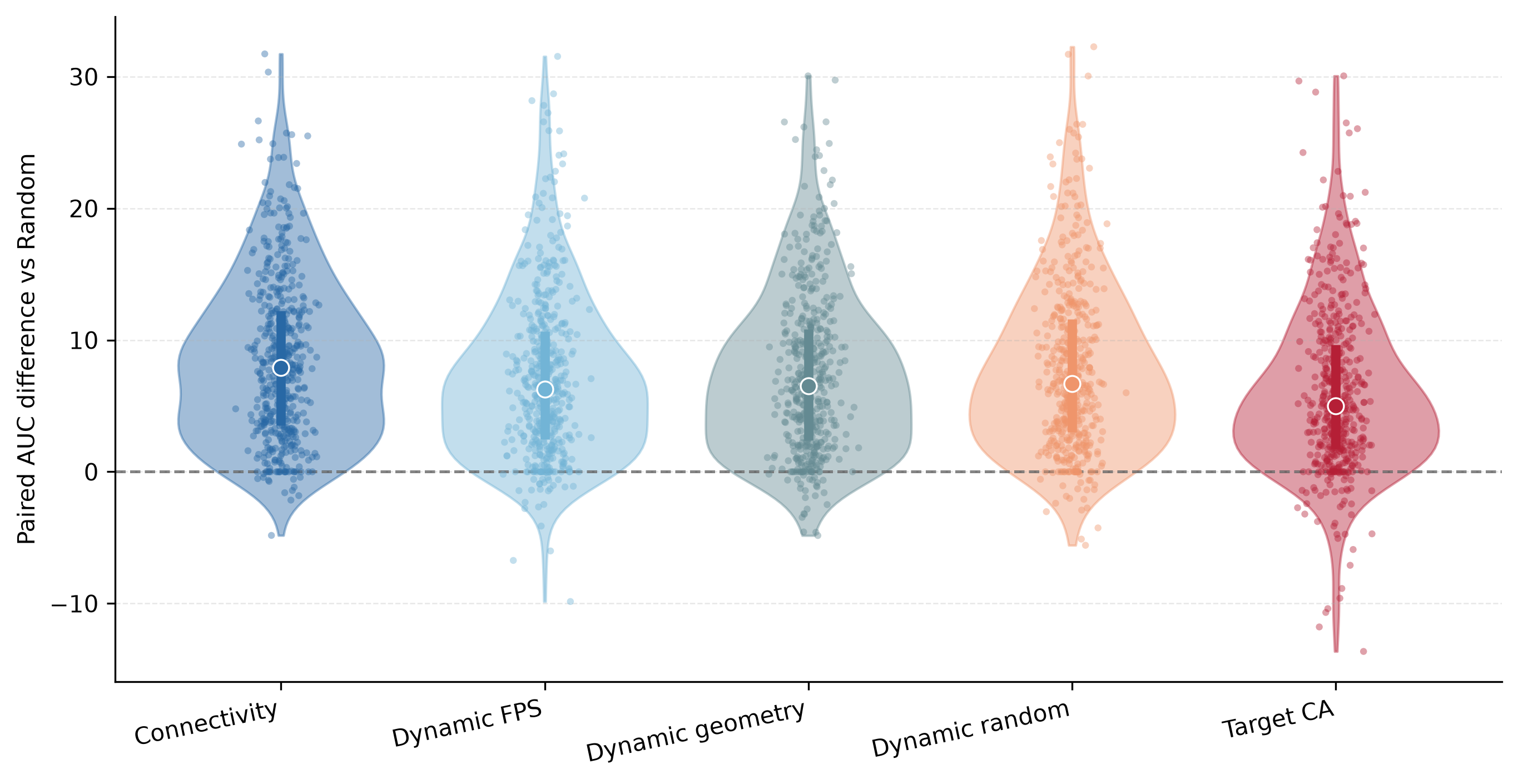
        }
        \caption{Paired AUC improvement over random.}
        \label{fig:e4_auc_violin}
    \end{subfigure}
    \hfill
    \begin{subfigure}[t]{0.325\textwidth}
        \centering
        \includegraphics[width=\linewidth]{
            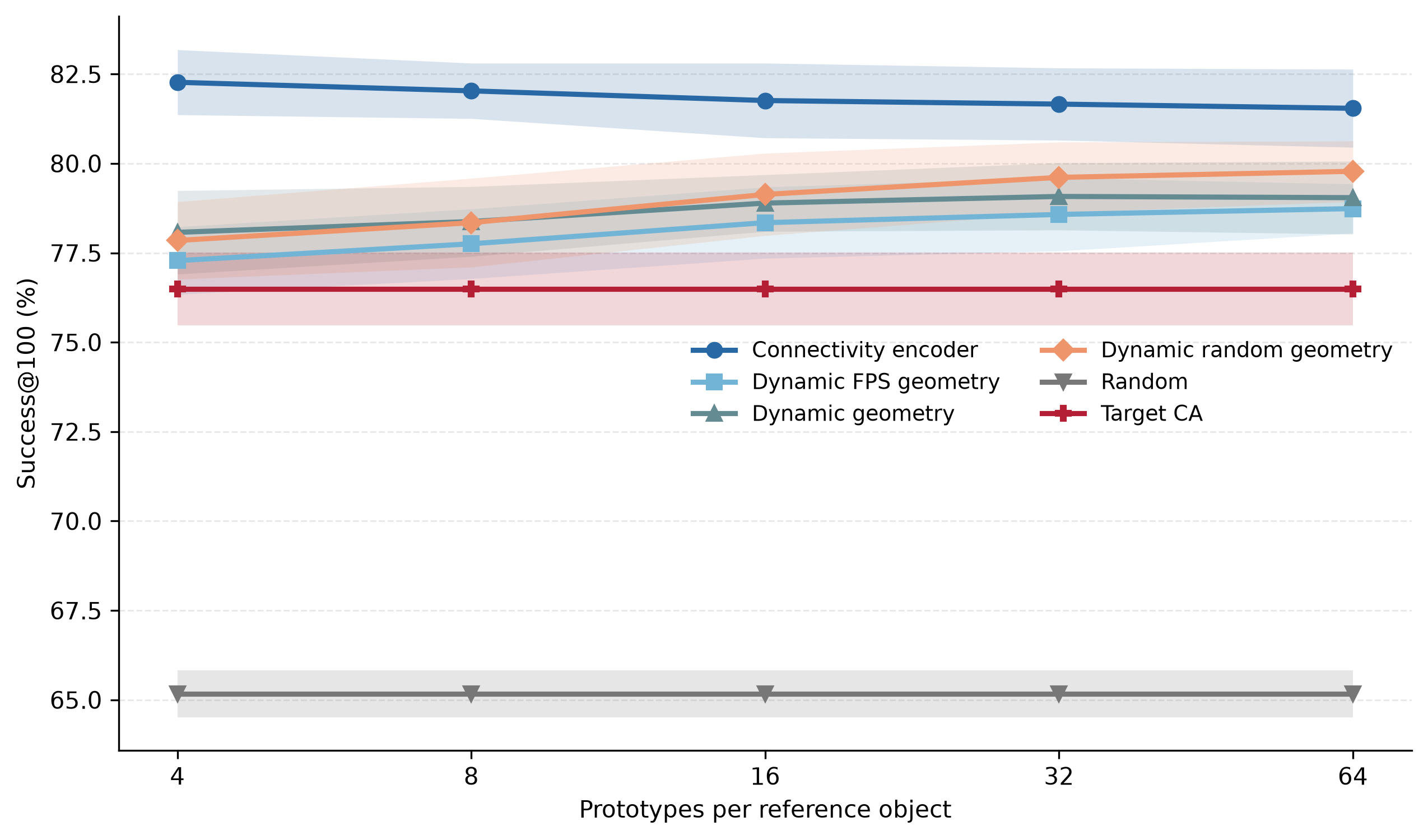
        }
        \caption{Prototype-capacity ablation.}
        \label{fig:e4_prototype_scaling}
    \end{subfigure}

    \caption{
        \textbf{E4: Cross-object grasp-space exploration.}
        (a) Success@100 as a function of the number of
        reference objects, with 16 prototypes per reference
        object.
        (b) Target-level paired differences in cumulative-success
        AUC relative to random exploration, for 16 prototypes
        per reference object and seed 0.
        (c) Success@100 as a function of prototype capacity
        with 50 reference objects.
        Shaded bands in (a) and (c) represent
        $\pm 1$ standard deviation across five seeds.
    }
    \label{fig:e4_combined}
\end{figure*}

\subsection{E3: Hidden-Viability Exploration and Exploitation}
\label{sec:e3_results}

Experiment~3 tests whether observations acquired under hidden grasp viability
improve subsequent discovery in a disjoint candidate pool, and whether this
benefit depends on the Phase~1 observation policy
(Table~\ref{tab:e3_multiscale}; Fig.~\ref{fig:e3_ca_exploitation}).

Under Random Phase~1 exploration, all informed exploitation rules substantially
outperformed Random ranking, with gains increasing with candidate-set size.
At $N=1600$, normalized Phase~2 AUC increased from $0.4995\pm0.0109$ for
Random to $0.7187\pm0.0892$ for nearest-success (NS),
$0.7614\pm0.0887$ for local viability (LV), and $0.7598\pm0.0845$ for
connectivity-aware (CA) exploitation. Thus, proximity to observed successes
provides a large benefit, while using local success/failure information provides most of the additional improvement beyond nearest-success ranking.

The same pattern was largely preserved following CA Phase~1 exploration.
At $N=1600$, NS, LV, and CA reached normalized AUCs of $0.7005$,
$0.7563$, and $0.7577$, respectively. Across candidate-set sizes, LV and CA
performed similarly, indicating that explicit multiscale connectivity provides
little additional predictive benefit beyond local success/failure geometry.

Finally, CA acquisition itself did not consistently improve downstream
exploitation: for a fixed exploitation rule, Random and CA Phase~1
observations produced similar performance, particularly at larger $N$.
Together, E2 and E3 therefore distinguish two roles of grasp-space structure:
explicit connectivity improves structural reconstruction efficiency, whereas
local viability geometry captures most of the additional benefit for
hidden-viability exploitation.

\begin{table*}[t]
\centering
\caption{E4 prototype-capacity ablation with 50 reference
objects. Success@100 and cumulative-success AUC are
reported as mean $\pm$ SD across five independently
trained seeds. Superscripts indicate Holm-adjusted
significance against Connectivity using two-sided paired
seed-level $t$-tests, with correction across five baseline
comparisons separately for each capacity and metric.
Bold indicates the highest mean within each
prototype-capacity column.}
\label{tab:e4_prototype_both_metrics}

{\scriptsize
\setlength{\tabcolsep}{9pt}
\renewcommand{\arraystretch}{1.15}

\begin{tabular}{lccccc}
\hline
\multicolumn{6}{c}{\textbf{(a) Success@100 (\%)}} \\
\hline
Method
& $P=4$ & $P=8$ & $P=16$ & $P=32$ & $P=64$ \\
\hline

Connectivity
& $\mathbf{82.27\pm0.91}$
& $\mathbf{82.03\pm0.77}$
& $\mathbf{81.76\pm1.04}$
& $\mathbf{81.66\pm1.01}$
& $\mathbf{81.54\pm1.09}$ \\

Dynamic FPS
& $77.28\pm0.94^{**}$
& $77.75\pm0.97^{**}$
& $78.34\pm1.00^{**}$
& $78.57\pm1.02^{**}$
& $78.74\pm0.69^{**}$ \\

Dynamic geometry
& $78.07\pm1.17^{**}$
& $78.38\pm0.98^{**}$
& $78.89\pm0.79^{**}$
& $79.08\pm0.94^{**}$
& $79.04\pm1.02^{**}$ \\

Dynamic random geometry
& $77.85\pm1.08^{**}$
& $78.34\pm1.24^{**}$
& $79.13\pm1.15^{**}$
& $79.61\pm0.99^{**}$
& $79.78\pm0.85^{**}$ \\

Target CA
& $76.49\pm1.02^{**}$
& $76.49\pm1.02^{**}$
& $76.49\pm1.02^{**}$
& $76.49\pm1.02^{**}$
& $76.49\pm1.02^{**}$ \\

Random
& $65.17\pm0.65^{**}$
& $65.17\pm0.65^{**}$
& $65.17\pm0.65^{**}$
& $65.17\pm0.65^{**}$
& $65.17\pm0.65^{**}$ \\

\hline
\multicolumn{6}{c}{\textbf{(b) Cumulative-success AUC}} \\
\hline
Method
& $P=4$ & $P=8$ & $P=16$ & $P=32$ & $P=64$ \\
\hline

Connectivity
& $\mathbf{41.64\pm0.45}$
& $\mathbf{41.50\pm0.39}$
& $\mathbf{41.43\pm0.49}$
& $\mathbf{41.36\pm0.48}$
& $\mathbf{41.31\pm0.53}$ \\

Dynamic FPS
& $39.01\pm0.48^{**}$
& $39.33\pm0.50^{**}$
& $39.65\pm0.48^{**}$
& $39.71\pm0.50^{**}$
& $39.78\pm0.36^{**}$ \\

Dynamic geometry
& $39.52\pm0.61^{**}$
& $39.71\pm0.39^{**}$
& $39.90\pm0.39^{**}$
& $40.01\pm0.45^{**}$
& $40.03\pm0.48^{**}$ \\

Dynamic random geometry
& $39.44\pm0.52^{**}$
& $39.69\pm0.63^{**}$
& $40.12\pm0.51^{**}$
& $40.33\pm0.44^{**}$
& $40.42\pm0.38^{**}$ \\

Target CA
& $38.74\pm0.51^{**}$
& $38.74\pm0.51^{**}$
& $38.74\pm0.51^{**}$
& $38.74\pm0.51^{**}$
& $38.74\pm0.51^{**}$ \\

Random
& $32.96\pm0.38^{**}$
& $32.96\pm0.38^{**}$
& $32.96\pm0.38^{**}$
& $32.96\pm0.38^{**}$
& $32.96\pm0.38^{**}$ \\

\hline
\end{tabular}
}

\vspace{0.3em}
\footnotesize
$P$ denotes the nominal maximum number of reference
prototypes per object; actual memory sizes may be smaller.
The number of reference objects is fixed at 50.
Each configuration is evaluated on 500 held-out target
objects per seed. Results are first averaged over target
objects and then summarized across five seeds.
Standard deviations quantify variation across seeds.
The marker $^{**}$ denotes $p_{\mathrm H}<0.01$.
All tests are two-sided and Holm-adjusted.
Exact adjusted $p$-values and confidence intervals
are reported in the appendix.
\end{table*}


\begin{table*}[t]
\centering
\caption{E4 reference-object scaling with $P=16$ nominal
prototypes per reference object. Success@100 and
cumulative-success AUC are reported as mean $\pm$ SD
across five independently trained seeds. Superscripts
indicate Holm-adjusted significance against Connectivity
using two-sided paired seed-level $t$-tests, with correction
across five baseline comparisons separately for each
reference count and metric. Bold indicates the highest
mean within each reference-count column.}
\label{tab:e4_reference_scaling}

\setlength{\tabcolsep}{3pt}
\renewcommand{\arraystretch}{1.15}
\resizebox{\textwidth}{!}{%
\begin{tabular}{lcccccccc}
\hline
\multicolumn{9}{c}{\textbf{(a) Success@100 (\%)}} \\
\hline
Method
& $N=6$ & $N=12$ & $N=25$ & $N=50$
& $N=100$ & $N=200$ & $N=400$ & $N=800$ \\
\hline

Connectivity
& $70.07\pm7.85$
& $\mathbf{77.21\pm8.21}$
& $\mathbf{81.20\pm0.89}$
& $\mathbf{81.76\pm1.04}$
& $\mathbf{81.82\pm1.35}$
& $\mathbf{82.27\pm0.92}$
& $\mathbf{82.65\pm0.77}$
& $\mathbf{82.79\pm0.80}$ \\

Dynamic FPS
& $68.52\pm6.21$
& $71.17\pm8.15$
& $77.27\pm0.69^{***}$
& $78.34\pm1.00^{***}$
& $79.06\pm0.92^{***}$
& $80.06\pm0.86^{***}$
& $80.72\pm0.96^{**}$
& $81.17\pm0.81^{***}$ \\

Dynamic geometry
& $70.50\pm5.99$
& $71.16\pm8.25$
& $77.26\pm0.51^{***}$
& $78.89\pm0.79^{***}$
& $79.88\pm1.01^{***}$
& $80.84\pm0.55^{**}$
& $81.12\pm0.74^{**}$
& $81.34\pm0.49^{**}$ \\

Dynamic random geometry
& $68.47\pm5.92$
& $70.99\pm8.02$
& $77.47\pm0.87^{***}$
& $79.13\pm1.15^{***}$
& $79.80\pm1.27^{***}$
& $80.82\pm0.95^{***}$
& $80.92\pm0.43^{**}$
& $81.25\pm0.50^{***}$ \\

Target CA
& $\mathbf{76.49\pm1.02}$
& $76.49\pm1.02$
& $76.49\pm1.02^{***}$
& $76.49\pm1.02^{***}$
& $76.49\pm1.02^{***}$
& $76.49\pm1.02^{***}$
& $76.49\pm1.02^{***}$
& $76.49\pm1.02^{***}$ \\

Random
& $65.17\pm0.65$
& $65.17\pm0.65$
& $65.17\pm0.65^{***}$
& $65.17\pm0.65^{***}$
& $65.17\pm0.65^{***}$
& $65.17\pm0.65^{***}$
& $65.17\pm0.65^{***}$
& $65.17\pm0.65^{***}$ \\

\hline
\multicolumn{9}{c}{\textbf{(b) Cumulative-success AUC}} \\
\hline
Method
& $N=6$ & $N=12$ & $N=25$ & $N=50$
& $N=100$ & $N=200$ & $N=400$ & $N=800$ \\
\hline

Connectivity
& $35.11\pm4.41$
& $\mathbf{39.01\pm4.34}$
& $\mathbf{41.10\pm0.39}$
& $\mathbf{41.43\pm0.49}$
& $\mathbf{41.44\pm0.68}$
& $\mathbf{41.64\pm0.49}$
& $\mathbf{41.90\pm0.37}$
& $\mathbf{41.93\pm0.36}$ \\

Dynamic FPS
& $34.39\pm3.35$
& $35.69\pm4.44$
& $38.96\pm0.31^{**}$
& $39.65\pm0.48^{**}$
& $39.98\pm0.42^{***}$
& $40.48\pm0.45^{***}$
& $40.89\pm0.49^{**}$
& $41.22\pm0.42^{***}$ \\

Dynamic geometry
& $35.40\pm3.24$
& $35.76\pm4.54$
& $39.00\pm0.18^{**}$
& $39.90\pm0.39^{**}$
& $40.47\pm0.49^{***}$
& $40.92\pm0.27^{**}$
& $41.11\pm0.41^{**}$
& $41.30\pm0.25^{**}$ \\

Dynamic random geometry
& $34.37\pm3.14$
& $35.63\pm4.39$
& $39.25\pm0.39^{***}$
& $40.12\pm0.51^{**}$
& $40.40\pm0.62^{***}$
& $40.91\pm0.48^{***}$
& $41.01\pm0.18^{**}$
& $41.19\pm0.22^{**}$ \\

Target CA
& $\mathbf{38.74\pm0.51}$
& $38.74\pm0.51$
& $38.74\pm0.51^{***}$
& $38.74\pm0.51^{***}$
& $38.74\pm0.51^{***}$
& $38.74\pm0.51^{***}$
& $38.74\pm0.51^{***}$
& $38.74\pm0.51^{***}$ \\

Random
& $32.96\pm0.38$
& $32.96\pm0.38$
& $32.96\pm0.38^{***}$
& $32.96\pm0.38^{***}$
& $32.96\pm0.38^{***}$
& $32.96\pm0.38^{***}$
& $32.96\pm0.38^{***}$
& $32.96\pm0.38^{***}$ \\

\hline
\end{tabular}%
}

\vspace{0.3em}
\footnotesize
$N$ denotes the number of reference objects; the
number of supervised training objects increases with $N$.
Each configuration is evaluated on 500 held-out target
objects per seed. Results are first averaged over target
objects and then summarized across five seeds.
Standard deviations quantify variation across seeds.
Significance markers indicate
$^{*}p_{\mathrm H}<0.05$,
$^{**}p_{\mathrm H}<0.01$, and
$^{***}p_{\mathrm H}<0.001$.
Unmarked comparisons are not significant at the
0.05 level. All tests are two-sided and Holm-adjusted.
\end{table*}

\subsection{E4: Connectivity-Aware Cross-Object Exploration}
\label{sec:e4_results}

Experiment~4 evaluates whether connectivity-informed reference
memories and structural retrieval improve grasp discovery on
held-out objects. We examine reference-object scaling,
target-level improvements, and prototype-memory capacity
(Fig.~\ref{fig:e4_combined};
Tables~\ref{tab:e4_reference_scaling}
and~\ref{tab:e4_prototype_both_metrics}).

\paragraph*{Reference-object scaling.}
The connectivity encoder consistently outperforms geometric
retrieval baselines for $N_{\mathrm{ref}}=25$--$800$
(Fig.~\ref{fig:e4_reference_scaling}), achieving the highest
mean Success@100 and cumulative-success AUC among all six
methods (Holm-adjusted $p<0.01$ against each geometric
baseline for both metrics).

At $N_{\mathrm{ref}}=25$ and $P=16$, connectivity achieves
$81.20\pm0.89$ successful grasps per 100 queries, compared
with $77.47\pm0.87$ for the strongest geometric baseline
(dynamic random geometry), a gain of $3.74$ successes
($p=0.0002$). AUC improves from $39.25\pm0.39$ to
$41.10\pm0.39$ ($p=0.0001$).
At $N_{\mathrm{ref}}=50$, connectivity achieves
$81.76\pm1.04$ successes and $41.43\pm0.49$ AUC,
exceeding dynamic geometry by $2.87$ successes and
$1.53$ AUC units ($p=0.0009$ and $p=0.0017$).
Because these methods share connectivity-selected
prototypes, this comparison isolates the contribution
of structural rather than pose-based retrieval.

With only 25 reference objects, connectivity achieves
approximately the performance of geometric methods using
200--400 references, suggesting an $8$--$16\times$
difference in reference-object requirements at comparable
observed performance. However, reference diversity and
total memory capacity also increase with reference count.
At $N_{\mathrm{ref}}=6$ and $12$, transfer is less reliable:
target-only connectivity performs best on average at six
references, and connectivity's advantages at 12 references
are not statistically significant. At $N_{\mathrm{ref}}=800$,
connectivity reaches $82.79\pm0.80$ successes and
$41.93\pm0.36$ AUC, although its advantage narrows as
geometric retrieval becomes more competitive.

\paragraph*{Target-level improvements.}
At $N_{\mathrm{ref}}=50$ and $P=16$, connectivity improves
cumulative-success AUC over random exploration by
$8.47$ units on average across five seeds
($41.43$ versus $32.96$).
The paired target-level distributions
(Fig.~\ref{fig:e4_auc_violin}) are predominantly positive
across exploration methods but overlap substantially and
include negative differences, indicating heterogeneous
benefits across objects. The violin plot shows one
representative seed; statistical inference uses paired
seed-level aggregates.

\paragraph*{Prototype-memory capacity.}
Varying prototypes per reference object from $P=4$ to
$64$ at fixed $N_{\mathrm{ref}}=50$ reveals limited
sensitivity of structural retrieval to memory capacity
(Fig.~\ref{fig:e4_prototype_scaling};
Table~\ref{tab:e4_prototype_both_metrics}).
Connectivity achieves $82.27\pm0.91$ successes and
$41.64\pm0.45$ AUC at $P=4$, versus
$81.54\pm1.09$ and $41.31\pm0.53$ at $P=64$.

Geometric baselines generally improve with capacity:
dynamic random geometry increases from
$77.85\pm1.08$ to $79.78\pm0.85$ successes.
Nevertheless, connectivity remains superior to all
five baselines at every capacity (Holm-adjusted
$p<0.01$ for both metrics), with its advantage over
dynamic geometry decreasing from $4.20$ to $2.50$
successes. This demonstrates robustness across a
16-fold range of nominal prototype capacities, without
implying an equivalent reduction in measured memory
consumption or computational cost.

Overall, E4 demonstrates that structural retrieval improves
cross-object grasp discovery beyond pose-based retrieval
using identical connectivity-selected prototypes.
The advantage is strongest with limited reference
experience, while geometric retrieval becomes more
competitive as reference collections and prototype
capacities increase.

\section{Discussion and Conclusion}
\label{sec:discussion_conclusion}

We investigated whether the organization of viable grasps
in $SE(3)$ can support exploration, prediction, and
cross-object transfer. Across four experiments, we found
that grasp-space structure is reproducible and useful,
although its benefits depend on the exploration objective.

E1 demonstrated reproducible, object-dependent multiscale
connectivity patterns. Connectivity signatures from
independent samples of the same object become increasingly
similar with sample size, while between-object differences
remain larger. This suggests that the observed structure
reflects object-specific properties rather than sampling
variability alone. Complementing previous work on grasp
manifolds and topological representations
\cite{pokorny2013grasp,hang2016evolution,
zarubin2012topological,mahler2018caging},
E1 establishes the empirical reproducibility of
multiscale connectivity in sampled viable grasp sets.

E2 showed that this structure can improve reconstruction
efficiency under known viability. Connectivity-aware sampling
approximated reference connectivity signatures with fewer
observations than random sampling and farthest-point sampling
(FPS), achieving up to a $4.76\times$ median sample-efficiency
improvement over random sampling at matched reconstruction
quality. Unlike FPS, which prioritizes geometric coverage,
connectivity-aware sampling targets structural features
such as component boundaries and potential connections.
Thus, geometric coverage alone is not necessarily sufficient
for efficient reconstruction of grasp-space organization.

E3 examined hidden viability, where successful and failed
observations inform predictions for untested candidates.
Both local-viability and connectivity-aware inference
outperformed nearest-success ranking but achieved similar
performance to each other. This suggests that local
success/failure geometry captures much of the information
relevant to predicting individual grasp outcomes.
The relative ordering remained largely unchanged under
connectivity-aware versus random observation acquisition.
Consequently, the advantages of explicit connectivity
for structural reconstruction in E2 do not directly
translate into improved hidden-viability exploitation.

E4 demonstrated that structural retrieval from
connectivity-selected reference memories improves sequential
grasp discovery on held-out objects. Local candidate-graph
signatures enable cross-object transfer without pointwise
pose correspondence or advance knowledge of target viability.
Compared with geometric retrieval using identical reference
prototypes, structural retrieval generally achieved higher
success rates, particularly with limited reference experience.
The advantage persisted across reference-set sizes and
prototype capacities, although geometric retrieval became
more competitive as reference resources increased.
Online evaluation time remained approximately constant
over the tested range, while offline training costs grew
with reference-set size.

Several limitations remain. First, connectivity-based
exploration represents discovered viable grasps but does
not directly incorporate failed grasps into its connectivity
graph. Joint modeling of viable and non-viable regions
could improve acquisition under limited observations.
Second, E4 uses handcrafted local structural signatures
and explicit reference memories; learned representations
and adaptive memory selection may improve scalability.
Furthermore, transfer was evaluated on held-out objects
without enforcing disjoint object categories, leaving
category-level generalization unestablished.
Finally, all experiments used precomputed candidate poses
and simulated grasp outcomes. Online grasp generation,
closed-loop simulation, and physical robotic evaluation
are necessary to assess robustness under perception
uncertainty and execution noise.

Overall, our findings distinguish three roles of grasp-space
structure: connectivity supports efficient reconstruction,
local success/failure geometry supports hidden-viability
prediction, and structural retrieval improves cross-object
transfer. These results motivate robotic exploration methods
that exploit relationships among candidate grasps while
adapting their structural representations to the task.

\clearpage
\bibliographystyle{IEEEtran}
\bibliography{references}
\clearpage
\appendices

\section{Construction and Interpretation of the Connectivity Signature}
\label{app:connectivity}

This appendix provides additional details on the multiscale connectivity
representation used throughout the experiments. The construction is based
on standard neighborhood-graph and zeroth-dimensional topological notions:
points are connected at a given scale according to pairwise distance, and
the evolution of connected components is examined as that scale varies
\cite{edelsbrunner2010computational}.

\section{Connectivity Graph}

For an object $O$, consider a finite set of successful grasp poses
\begin{equation}
\mathcal{G}^{+}_{N}(O)
=
{g_1,\ldots,g_N},
\qquad
g_i\in SE(3).
\end{equation}

The connectivity analysis is performed over successful grasps only.
Failed grasps therefore do not appear as vertices of the viable-set
connectivity graph.

For every pair of successful grasps $(g_i,g_j)$, we compute the normalized
$SE(3)$ distance $d(g_i,g_j)$ defined in
Section~\ref{sec:se3_metric}. For a neighborhood scale $\epsilon$, we
construct an undirected $\epsilon$-neighborhood graph
\begin{equation}
G_{\epsilon}
=
(V,E_{\epsilon}),
\end{equation}
where
\begin{equation}
V=\mathcal{G}^{+}_{N}(O)
\end{equation}
and
\begin{equation}
(g_i,g_j)\in E_{\epsilon}
\quad\Longleftrightarrow\quad
d(g_i,g_j)\leq\epsilon.
\end{equation}

This construction corresponds to the graph underlying a
Vietoris--Rips-type filtration: vertices represent sampled points and
edges are introduced when their pairwise distance falls below a specified
scale \cite{edelsbrunner2010computational}. In the present work, however,
we use only the resulting graph connectivity and do not construct or
analyze higher-dimensional simplices.

Two successful grasps are therefore directly adjacent when their distance
does not exceed $\epsilon$. They may also belong to the same connected
component without being directly adjacent if there exists a sequence of
intermediate successful grasps connecting them.

\section{Connectivity Statistics}

At each scale $\epsilon$, we summarize $G_{\epsilon}$ using two
connectivity statistics. The first is the zeroth Betti number
\begin{equation}
\beta_0(\epsilon),
\end{equation}
which equals the number of connected components at that scale
\cite{edelsbrunner2010computational}. Because the number of components
depends directly on the number of sampled grasps, we use the normalized
component count
\begin{equation}
B(\epsilon)
=
\frac{\beta_0(\epsilon)}{N}.
\end{equation}

Large values of $B(\epsilon)$ indicate a highly fragmented viable grasp
set at that scale, whereas small values indicate that the successful
grasps have merged into relatively few connected regions.

The second statistic describes the size of the dominant connected region.
Let $C_{\max}(\epsilon)$ denote the largest connected component of
$G_{\epsilon}$. We define
\begin{equation}
L(\epsilon)
=
\frac{|C_{\max}(\epsilon)|}{N}.
\end{equation}

Thus, $L(\epsilon)$ is the fraction of sampled successful grasps contained
in the largest connected component. The two quantities capture
complementary properties of the viable set: $B(\epsilon)$ measures its
fragmentation, while $L(\epsilon)$ measures the dominance of its largest
connected region.

\section{Multiscale Connectivity Signature}

Connectivity is inherently scale dependent. At small values of
$\epsilon$, most grasps are isolated or belong to small connected
components. As $\epsilon$ increases, additional edges appear and
components progressively merge. Consequently, $B(\epsilon)$ decreases,
while $L(\epsilon)$ increases.

We evaluate these two statistics at a fixed sequence of $K$ neighborhood
scales,
\begin{equation}
    \epsilon_1 < \epsilon_2 < \cdots < \epsilon_K.
\end{equation}

For a viable grasp set $\mathcal{G}^{+}_{N}$, we collect the normalized
component counts across these scales into the vector
\begin{equation}
    \mathbf{B}
    =
    \left(
        B(\epsilon_1),
        B(\epsilon_2),
        \ldots,
        B(\epsilon_K)
    \right),
\end{equation}
and the corresponding largest-component fractions into
\begin{equation}
    \mathbf{L}
    =
    \left(
        L(\epsilon_1),
        L(\epsilon_2),
        \ldots,
        L(\epsilon_K)
    \right).
\end{equation}

We define the multiscale connectivity signature of the viable grasp set
as the pair
\begin{equation}
    T(\mathcal{G}^{+}_{N})
    =
    \left(
        \mathbf{B},
        \mathbf{L}
    \right).
    \label{eq:connectivity_signature}
\end{equation}

Thus, the connectivity signature consists of two curves sampled at the
same $K$ neighborhood scales: $\mathbf{B}$ describes how rapidly the
viable grasp set changes from many disconnected components to fewer
connected regions, while $\mathbf{L}$ describes how rapidly a dominant
connected component emerges.
\section{Cross-Object Grasp Exploration}
\subsection{Implementation and Mathematical Details}
\label{app:e4_math}

This appendix specifies the pose metric,
reference-prototype construction, structural
signatures, retrieval distances, and neural
architecture used in Experiment~4.
Training and evaluation procedures are given
in Algorithms~\ref{alg:e4_training}
and~\ref{alg:e4_evaluation}.

\subsubsection{Pose Preprocessing and Reference Memory}
\label{app:e4_pose}

Unlike the normalized geodesic metric in
Section~\ref{sec:se3_metric}, E4 uses a
weighted Euclidean approximation in
translation and rotation-vector coordinates.
This avoids repeated geodesic rotation
computations during nearest-neighbor retrieval
and reference graph construction.

For each object, candidate translations
are centered at their coordinatewise median
and divided by the median Euclidean norm
of the centered translations. Rotations
are represented by rotation vectors.
For poses $x=(t_x,r_x)$ and $z=(t_z,r_z)$,
the distance is
\begin{equation}
d_\rho(x,z)=
\left(
\|t_x-t_z\|_2^2+
\rho^2\|r_x-r_z\|_2^2
\right)^{1/2},
\qquad \rho=0.5.
\label{eq:e4_distance}
\end{equation}
The fixed coefficient $\rho=0.5$
reduces the relative contribution of
rotational differences and was not
separately optimized.

The neural predictor uses
$\phi(x)=[t_x,\sin r_x,\cos r_x]
\in\mathbb R^9$, with trigonometric
functions applied componentwise.

For each reference object, a proximity
graph is constructed over successful
grasps, with edges between pairs satisfying
$d_\rho(x,z)\leq\varepsilon$,
where $\varepsilon=0.5$.
Connected components containing fewer
than two successful grasps are discarded.

In contrast to the global farthest-point
sampling used in Experiment~2, E4 selects
prototypes separately within each retained
component. The first prototype of
component $C_c$ is its medoid:
\begin{equation}
p_{c,1}=
\arg\min_{x\in C_c}
\sum_{z\in C_c}d_\rho(x,z).
\label{eq:e4_medoid}
\end{equation}
Subsequent prototypes are selected by
farthest-point sampling
~\cite{gonzalez1985clustering}:
\begin{equation}
p_{c,j}=
\arg\max_{x\in C_c}
\min_{p\in\mathcal P_c^{(j-1)}}
d_\rho(x,p),
\label{eq:e4_fps}
\end{equation}
where $\mathcal P_c^{(j-1)}$ contains
the previously selected prototypes
from component $C_c$.

The per-object budget $P$ is allocated
across retained components in successive
rounds, with at most eight prototypes
per component. This preserves representation
of multiple successful-grasp regions
without allowing a single large component
to consume the entire memory budget.

Let $M_o$ denote the number of retained
components of reference object $o$,
and $K_o$ its actual prototype count.
Then
\begin{equation}
K_o\leq
\min\left\{
P,\sum_{c=1}^{M_o}
\min(8,|C_{o,c}|)
\right\}.
\label{eq:e4_object_capacity}
\end{equation}
Consequently, the actual memory size
may be smaller than the nominal budget,
particularly for objects containing
few connected components.

\subsubsection{Structural Signatures}
\label{app:e4_signatures}

The connectivity encoder represents
local candidate-space geometry using
structural signatures computed from
unlabeled poses.

Let $X=\{x_i\}_{i=1}^{n}$ be the
complete candidate-pose set of an object.
For each $x_i$, let $N_i$ contain its
$k_s=8$ nearest neighbors under $d_\rho$,
excluding $x_i$, and define
$d_{ij}=d_\rho(x_i,x_j)$.
The fixed neighborhood size captures
local geometric organization while
limiting the influence of distant poses.

For each reference, validation, or test
object, a distance scale is computed
independently from its unlabeled
candidate set:
\begin{equation}
\begin{aligned}
a&=\max\left\{
\operatorname{median}_i
\max_{j\in N_i}d_{ij},
10^{-6}
\right\},\\
u_{ij}&=\frac{d_{ij}}{a}.
\end{aligned}
\label{eq:e4_scale}
\end{equation}
This normalization reduces differences
in local distance scales across objects.

The structural signature of candidate
$x_i$ is
\begin{equation}
\begin{aligned}
s_i=\bigg[
&\overline u_i,\;
\operatorname{sd}(u_i),\;
\max_{j\in N_i}u_{ij},\\
&\frac{1}{k_s}
\sum_{j\in N_i}
\mathbf 1[u_{ij}\leq1],\\
&v_i,\;m_i,\;c_i,\;v_i
\bigg],
\end{aligned}
\label{eq:e4_signature}
\end{equation}
where $\overline u_i$ and
$\operatorname{sd}(u_i)$ are the mean
and standard deviation of normalized
neighbor distances. The fourth component
measures the fraction of neighbors
within the object-specific scale $a$.

The remaining descriptors are
\begin{equation}
\begin{aligned}
v_i&=\frac{1}{k_s}
\sum_{j\in N_i}
\frac{d_{j,(k_s)}}{a},\\
m_i&=\frac{1}{k_s}
\sum_{j\in N_i}
\mathbf 1[i\in N_j],\\
c_i&=\frac{1}{k_s^2}
\sum_{j\in N_i}
|N_j\cap N_i|.
\end{aligned}
\label{eq:e4_graph_features}
\end{equation}
Here $d_{j,(k_s)}$ is the distance
from $x_j$ to its $k_s$th nearest
neighbor. The descriptors capture
complementary properties of local
candidate-space geometry:
$v_i$ measures neighborhood scale,
$m_i$ measures reciprocal
nearest-neighbor relationships,
and $c_i$ measures shared-neighbor
overlap. These properties are related
to established nearest-neighbor and
graph-based similarity methods
~\cite{levina2005maximum,
jarvis1973clustering,ertoz2003finding}.

The evaluated implementation uses
eight signature channels, with $v_i$
appearing in both the fifth and eighth
positions.

\subsubsection{Structural Retrieval}
\label{app:e4_structural_retrieval}

Each retained reference prototype
is assigned the structural signature
of its nearest candidate in the
corresponding source object's complete
unlabeled candidate-pose set.

To prevent components with larger
numerical variation from dominating
structural distances, featurewise
normalization factors are computed
from the retained reference signatures:
\begin{equation}
\sigma_\ell=
\max\left\{
\operatorname{sd}_i(s_{i\ell}^{\rm ref}),
0.05
\right\}.
\label{eq:e4_signature_scale}
\end{equation}
The minimum scale of $0.05$ limits
the influence of nearly constant
components and prevents division
by zero.

For a target candidate $x$ and
reference prototype $p_i$, the
normalized structural distance is
\begin{equation}
d_s(x,p_i)=
\left\|
\frac{s(x;X)-s_i^{\rm ref}}{\sigma}
\right\|_2,
\label{eq:e4_structdist}
\end{equation}
where division is elementwise.
Smaller distances indicate greater
similarity between local candidate-space
neighborhoods.

\subsection{E4: Training and Evaluation Algorithms}
\label{app:e4_algorithm}

\begin{algorithm}[H]
\caption{E4: Reference Memory Construction and Encoder Training}
\label{alg:e4_training}
\footnotesize
\begin{algorithmic}[1]

\Require Reference/training objects
$\mathcal D_{\rm ref}=\mathcal D_{\rm train}$;
disjoint validation objects $\mathcal D_{\rm val}$;
prototype limit $P$; seeds $\mathcal S$

\Ensure Trained encoders and reference memories

\For{each seed $s\in\mathcal S$}

    \State $\mathcal M_{\rm conn}\gets\varnothing$

    \For{each reference object
    $o\in\mathcal D_{\rm ref}$}

        \State Identify successful grasps $\mathcal G_o^+$

        \State Construct $\varepsilon$-neighborhood
        graph using $d_\rho$

        \State Retain components with at least two grasps

        \State Select up to $P$ prototypes via medoids and FPS

        \State Store prototypes, object IDs,
        and component IDs in $\mathcal M_{\rm conn}$

    \EndFor

    \State Build count-matched random and FPS memories

    \State Compute structural signatures from
    unlabeled candidate-pose graphs

    \State Associate signatures with reference prototypes

    \For{each learned method
    $m\in\{\mathrm{RandGeom},\mathrm{FPSGeom},
    \mathrm{Geom},\mathrm{Conn}\}$}

        \State Initialize encoder $\theta_m^{(s)}$

        \State Select method-specific reference memory

        \State Use geometric retrieval for
        RandGeom, FPSGeom, and Geom;
        structural retrieval for Conn

        \State Apply outcome-conditioned structural
        reranking for Conn

        \If{dynamic geometry is enabled}
            \State Apply outcome-conditioned geometric
            reranking to geometric variants
        \EndIf

        \State Train using sampled observation
        histories and binary cross-entropy

        \State Exclude own-object reference prototypes

        \State Select checkpoint using
        held-out validation objects

    \EndFor

\EndFor

\end{algorithmic}
\end{algorithm}

\begin{algorithm}[H]
\caption{E4: Sequential Cross-Object Grasp Exploration}
\label{alg:e4_evaluation}
\footnotesize
\begin{algorithmic}[1]

\Require Encoders $\theta_m^{(s)}$;
memories $\mathcal M_m^{(s)}$;
test pool $\mathcal D_{\rm test}$;
budget $B$; seeds $\mathcal S$

\Ensure Success-at-budget and cumulative-success AUC

\For{each seed $s\in\mathcal S$}

    \State Sample disjoint test objects
    $\mathcal D_{\rm test}^{(s)}$

    \For{each object $o\in\mathcal D_{\rm test}^{(s)}$}

        \State Obtain unlabeled candidates $\mathcal X_o$

        \For{$m\in\{\mathrm{Random},\mathrm{TargetCA},
    \mathrm{RandGeom},$}
    \Statex \hspace{\algorithmicindent}
    $\mathrm{FPSGeom},\mathrm{Geom},\mathrm{Conn}\}$
            \State Initialize $H_0=(\varnothing,\varnothing)$,
            $U_0=\mathcal X_o$, $C_0=0$

            \For{$t=0,\ldots,\min(B,|\mathcal X_o|)-1$}

                \If{$m=\mathrm{Random}$}
                    \State Sample $x_t$ uniformly from $U_t$

                \ElsIf{$m=\mathrm{TargetCA}$}
                    \State Select $x_t$ via target-only
                    connectivity-aware exploration

                \Else
                    \State Retrieve reference neighbors;
                    rerank using $H_t$ for dynamic methods

                    \State $x_t\gets
                    \arg\max_{x\in U_t}
                    \widehat P_{\theta_m^{(s)}}
                    (y=1\mid x,H_t,\mathcal M_m^{(s)})$
                \EndIf

                \State Query $x_t$; observe $y_t\in\{0,1\}$

                \State $H_{t+1}\gets H_t\cup\{(x_t,y_t)\}$

                \State $U_{t+1}\gets U_t\setminus\{x_t\}$;
                $C_{t+1}\gets C_t+y_t$

            \EndFor

            \State Store trajectory $\{C_t^{(m,s,o)}\}$

        \EndFor
    \EndFor

    \State Compute seed-level success at $B$
    and cumulative-success AUC

\EndFor

\State Report mean $\pm$ SD across seed-level
target means, prototype counts, and computational costs

\end{algorithmic}
\end{algorithm}

\subsection{E4: Computational Cost and Reference-Memory Scaling}
\label{app:e4_computational_cost}

We evaluate computational cost across
$N_{\mathrm{ref}}$
reference objects, with at most 16 prototypes per object.
Results are averaged over five seeds, with 500 evaluation
objects and up to 100 exploration steps per object.
Experiments use Apple Silicon with PyTorch MPS acceleration.

Table~\ref{tab:e4_computational_cost} summarizes representative
reference-set sizes for the connectivity and dynamic geometry
encoders. Online time includes end-to-end evaluation per
exploration step, rather than retrieval alone. Offline costs
include neural-network training and shared reference-memory
construction.

\begin{table}[!t]
\centering
\caption{Computational cost across reference-set sizes.
Online evaluation (ms/step), offline training (s),
and memory construction (s). C: connectivity;
G: dynamic geometry. Mean $\pm$ SD over five seeds.}
\label{tab:e4_computational_cost}
\scriptsize
\setlength{\tabcolsep}{2pt}
\renewcommand{\arraystretch}{1.15}
\begin{tabular}{rccccc}
\toprule
& \multicolumn{2}{c}{Online (ms/step)}
& \multicolumn{2}{c}{Training (s)}
& Memory \\
\cmidrule(lr){2-3}
\cmidrule(lr){4-5}
$N_{\rm ref}$ & C & G & C & G & Build (s) \\
\midrule
6 &
$9.25\pm0.26$ & $8.30\pm0.25$ &
$3.6\pm0.5$ & $1.1\pm0.5$ &
$0.07\pm0.01$ \\
50 &
$9.67\pm0.25$ & $8.46\pm0.28$ &
$19.3\pm1.0$ & $8.8\pm1.3$ &
$0.25\pm0.03$ \\
200 &
$9.85\pm0.29$ & $8.56\pm0.32$ &
$70.8\pm4.8$ & $41.1\pm4.5$ &
$1.00\pm0.05$ \\
800 &
$9.41\pm0.22$ & $8.18\pm0.19$ &
$334.7\pm15.4$ & $199.0\pm22.8$ &
$4.00\pm0.17$ \\
\bottomrule
\end{tabular}
\end{table}
Online evaluation time remains approximately constant as
the reference set grows from 6 to 800 objects:
$9.25$--$9.41$ ms/step for connectivity retrieval,
compared with $8.30$--$8.18$ ms/step for dynamic geometry.
At 800 references, connectivity retrieval incurs approximately
15\% greater online cost. In contrast, connectivity-encoder
training time increases from $3.64$ to $334.66$ s, while
shared memory construction increases from $0.07$ to $4.00$ s.

These results indicate that reference-set growth primarily
affects offline computation, with limited impact on online
evaluation time over the tested range. Because training-set
size and reference-memory capacity increase jointly, these
measurements characterize their combined scaling rather than
the isolated effect of memory capacity.

\clearpage
\section*{Code and Data Availability}

Code for all experiments and analyses in this study is publicly available at
\url{https://github.com/maksimkazanskii/connectivity_grasping}.
The repository contains the implementations of the connectivity analysis,
connectivity-aware exploration, hidden-viability experiments, cross-object
retrieval, and scripts used to generate the reported results and figures.
The experiments use the publicly available ACRONYM grasp dataset.

\section*{Generative AI Statement}

ChatGPT (GPT-5.6 Sol, OpenAI) was used to assist with language editing,
code development, and debugging. All experimental design, analysis,
interpretation, and scientific conclusions were developed and verified
by the author. All AI-assisted content was reviewed by the author, who
takes full responsibility for the manuscript.

\end{document}